\documentclass[10pt,journal]{IEEEtran}

\ifCLASSOPTIONcompsoc
  \usepackage[nocompress,noadjust]{cite}
\else
  \usepackage{cite}
\fi

\ifCLASSINFOpdf
  \usepackage[pdftex]{graphicx}
  \graphicspath{{./pdf/}{./jpeg/}{./graphics/}{./ME/Figs/}{./photos/}}
\else
\fi

\usepackage{amsmath}
\usepackage{lipsum}
\usepackage{mathtools}
\usepackage{cuted}
\ifCLASSOPTIONcompsoc
  \usepackage[caption=false,font=footnotesize,labelfont=sf,textfont=sf]{subfig}
\else
  \usepackage[caption=false,font=footnotesize]{subfig}
\fi
\usepackage{ragged2e}

\usepackage{tikz}
\usetikzlibrary{shapes,arrows,scopes,positioning,calc,math,patterns,decorations.pathmorphing,decorations.markings,decorations.pathreplacing,spy,angles,quotes}
\usepackage{ctable}
\makeatletter
\def\thickhline{%
  \noalign{\ifnum0=`}\fi\hrule \@height \thickarrayrulewidth \futurelet
   \reserved@a\@xthickhline}
\def\@xthickhline{\ifx\reserved@a\thickhline
               \vskip\doublerulesep
               \vskip-\thickarrayrulewidth
             \fi
      \ifnum0=`{\fi}}

\usepackage{array}

\pretocmd\@bibitem{\color{black}\csname keycolor#1\endcsname}{}{\fail}
\newcommand\citecolor[1]{\@namedef{keycolor#1}{\color{red}}}
\makeatother

\newlength{\thickarrayrulewidth}
\usepackage{multirow}
\usepackage{mcode}

\usepackage[inline]{asymptote}

\usepackage{siunitx}
\usepackage{nicefrac}

\let\oldfrac\frac% Store \frac
\makeatletter
\newcommand{\groupit}[1]{(#1)}% To group...
\newcommand{\nogroupit}[1]{#1}% ...or not to group
\renewcommand{\frac}[2]{%
  \setbox\z@\hbox{$#1$}% Store numerator
  \setbox\tw@\hbox{$#2$}% Store denominator
  \ifdim\wd\z@>1em \let\groupornot@i\groupit\else\let\groupornot@i\nogroupit\fi% Measure numerator
  \ifdim\wd\tw@>1em \let\groupornot@ii\groupit\else\let\groupornot@ii\nogroupit\fi% Measure denominator
  \mathchoice
    {\oldfrac{#1}{#2}}% display style
    {\groupornot@i{#1}/\groupornot@ii{#2}}% text style
    {\groupornot@i{#1}/\groupornot@ii{#2}}% script style
    {\groupornot@i{#1}/\groupornot@ii{#2}}% script-script style
}
\usepackage{xcolor}
\usepackage[normalem]{ulem}
\usepackage{soul}
\usepackage{lscape}
\usepackage[utf8]{inputenc}
\usepackage{float}
\usepackage{rotating}
\usepackage{pifont}
\newcommand{\cmark}{\ding{51}}%
\newcommand{\xmark}{\ding{55}}%

\begin{document}
\title{A Survey on Hybrid Motion Planning Methods for Automated Driving Systems}

\author{MReza~Alipour~Sormoli$^{1}$, Konstantinos Koufos$^{1}$,
Mehrdad~Dianati$^{1,2}$,~\IEEEmembership{Senior~Member,~IEEE,}
and Roger Woodman$^{1}$%
%\thanks{This research is sponsored by the Centre for Doctoral Training to Advance the Deployment of Future Mobility Technologies at the University of Warwick.}%
\thanks{$^{1}$ WMG, University of Warwick, Coventry, CV4 7AL, UK. {\tt\footnotesize mreza.alipour@warwick.ac.uk}
}
\thanks{$^{2}$ School of Electronics, Electrical Engineering and Computer Science, Queen’s University of Belfast, UK. {\tt\footnotesize m.dianati@qub.ac.uk}}
% \thanks{$^{\#}$Corresponding author: \texttt{mreza.alipour@warwick.ac.uk}}
% \tt\smal
}

\markboth{-}{}%
%{Shell \MakeLowercase{\textit{et al.}}: Bare Demo of IEEEtran.cls for IEEE Journals}

\maketitle

\begin{abstract}
Motion planning is an essential element of the modular architecture of autonomous vehicles, serving as a bridge between upstream perception modules and downstream low-level control signals.
Traditional motion planners were initially designed for specific Automated Driving Functions (ADFs), yet the evolving landscape of highly automated driving systems (ADS) requires motion for a wide range of ADFs, including unforeseen ones. This need has motivated the development of the ``hybrid" approach in the literature, seeking to enhance motion planning performance by combining diverse techniques, such as data-driven (learning-based) and logic-driven (analytic) methodologies.
Recent research endeavours have significantly contributed to the development of more efficient, accurate, and safe hybrid methods for Tactical Decision Making (TDM) and Trajectory Generation (TG), as well as integrating these algorithms into the motion planning module. Owing to the extensive variety and potential of hybrid methods, a timely and comprehensive review of the current literature is undertaken in this survey article.
We classify the \textit{hybrid} motion planners based on the types of components they incorporate, such as combinations of sampling-based with optimization-based/learning-based motion planners. The comparison of different classes is conducted by evaluating the addressed challenges and limitations, as well as assessing whether they focus on TG and/or TDM. We hope this approach will enable the researchers in this field to gain in-depth insights into the identification of current trends in hybrid motion planning and shed light on promising areas for future research.

\end{abstract}

% Note that keywords are not normally used for peerreview papers.

\IEEEpeerreviewmaketitle

\section{Introduction}

\IEEEPARstart{D}{eveloping} highly automated driving systems (ADS) promise to transform the automotive industry in the coming decades. Besides improving travel efficiency and passenger comfort, ADS can help improve their safety by reducing the number of accidents, which cause over 1.3 million deaths and even more permanent life-changing injuries annually \cite{world2018global}. Despite remarkable progress since the early 1980s \cite{dickmanns1988dynamic}, a significant boost in autonomous driving progress occurred through the Defense Advanced Research Projects Agency's (DARPA) competitions in 2004 and 2007. The Urban Challenge, in particular, marked a pivotal moment highlighting the significance of autonomous vehicles (AVs)\cite{buehler20072005,buehler2009darpa}. The performance of the vehicles in these competitions proved the feasibility of the automated driving concept, however many challenges before the mass commercialisation of the technology remained unaddressed. 

The two main approaches for developing ADS are the end-to-end and the modular solution~\cite{Teng2023}. The former directly maps the sensor data to control signals for the vehicle actuators whereas, in the latter, the problem is broken down into simpler sub-problems such as perception, localisation, behaviour prediction of other road users, and motion planning. In the modular approach, which is the focus of the current survey, a great number of challenges lie in the design and validation of motion planning and control. This module utilises the semantic knowledge of the surrounding environment generated by other modules, such as perception and prediction, and has to decide and calculate ``how" to move the ADS in different situations or driving scenarios.
% Therefore, its efficacy and performance are naturally affected by the quality of the perception and prediction algorithms. Nevertheless, even with perfect sensing, perception, and prediction, the motion planning module has to decide and calculate ``how" to move the ADS in different situations or driving scenarios. 
Two functions of trajectory generation (TG) and tactical decision-making (TDM) are the backbone of motion planning and have been the focal point of research~\cite{gonzalez2015review,ulbrich2015towards}. 
%Overcoming the open challenges in designing these functions can enable more reliable and safer driving control in ADS.
\begin{figure*}[t]
\centering
\includegraphics[width=.9\linewidth]{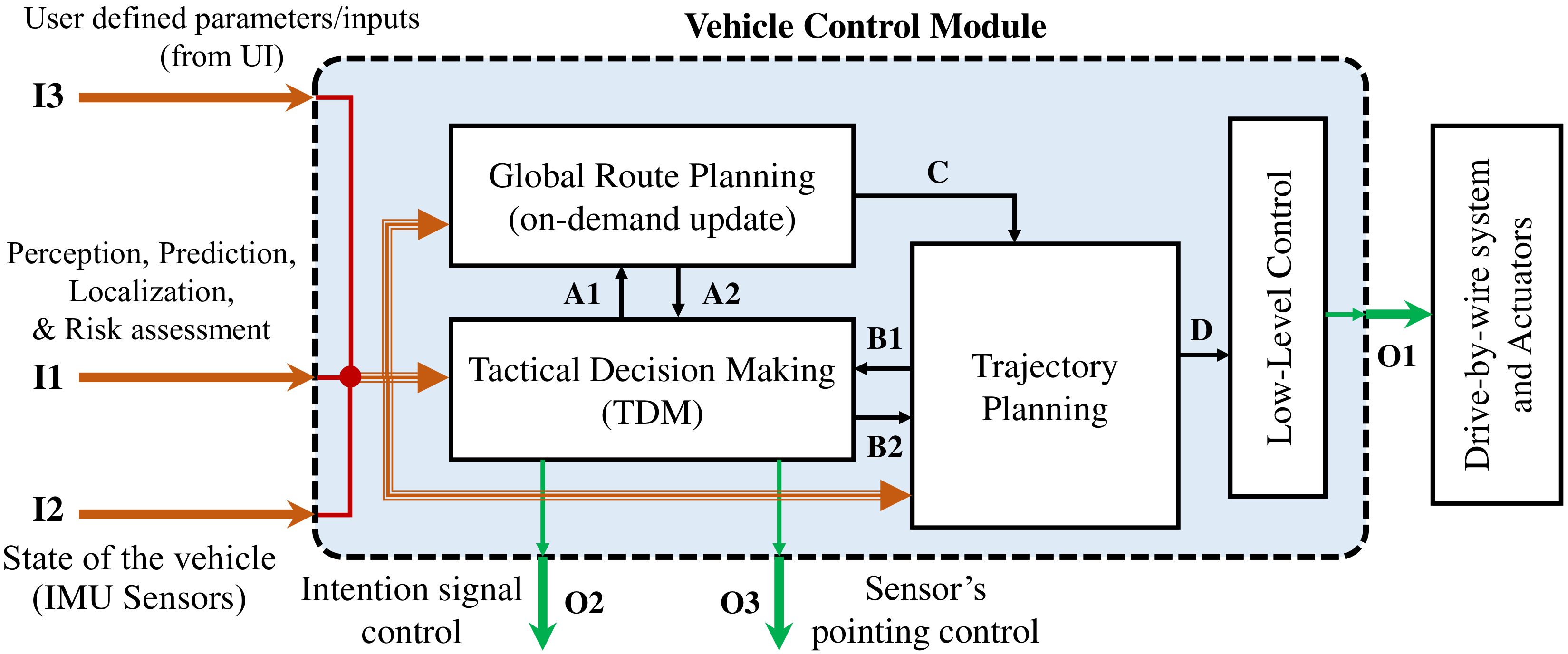}
\caption{The overall architecture of the control module for AVs in the modular architecture. The input and output signals are shown by \textbf{I} and \textbf{O} in brown and green colours, respectively. The functionality of the signals is explained in Section~\ref{bckg}.}
\label{fig:cont_layout}
\centering
\end{figure*}

The function of motion planning is to generate the best trajectory in terms of safety, comfort, and travel efficiency within a finite, deterministic, and short time period (high update frequency) to ensure \textit{real-time} computation. 
%The TG and TDM functions interactively operate to plan the motion in a short, mid, or long-range time horizon. 
In complex road environments, where a great variety of participants with different behaviours share the same space, Autonomous Vehicles (AVs) should be able to react properly and in real-time, which increases the demands and constraints put on the motion planning module. During the last decade, a large number of studies have been devoted to address several research questions in motion planning, which has led to numerous reviews and survey papers being published, classifying and categorizing various methods utilized in the AV design, along with their pros and cons~\cite{gonzalez2015review, katrakazas2015real,Long2023}. The performances of different methods have been evaluated in miscellaneous driving scenarios, including structured environments, such as highways~\cite{claussmann2019review}, or unstructured environments (e.g., parking lots). Moreover, following the recent progresses in data-driven algorithms and cutting-edge computational resources, Wilko et al. in~\cite{schwarting2018planning}, and recently, Teng et al. in~\cite{Teng2023} reviewed the decision-making and planning methods in AVs with a focus on learning-based algorithms, such as the end-to-end approach. However, as motion control for AVs is a very active research topic with many open research questions, recent studies have also been focusing on a new paradigm, namely, the \textbf{\textit{hybrid}} motion planning framework. 

%The hybrid approach aspires to improve the overall performance of motion planning using a combination of traditional motion planning methods, each separately restricted to particular situations, manoeuvres, or missions. The hybrid motion planning framework aims to enhance the performance (e.g., higher accuracy at a low computational cost) for a broad range of driving contexts by appropriately combining various techniques.
The hybrid approach aspires to improve the overall performance of motion planning using a combination of traditional motion planning techniques. For example, an optimization-based method could be used to create a dataset that maps the driving context, including perception and prediction, onto optimal trajectories. Then, a data-driven method can learn this relationship provided that a sufficient amount of training data has been generated. In this case, the resulting hybrid motion planner leverages both the real-time inference properties of data-driven methods and the optimal solutions attained by optimization-based techniques. Another example is the initialization of an optimization-based technique with a feasible trajectory obtained via a sampling-based method. In that case, the resulting hybrid method relaxes the high computational demands of optimization techniques because the search space is limited in the neighbourhood of the best sample. 
%An extensive review and categorization of hybrid motion planning methods in the literature is the main focus of this article.   

Due to the diversity of motion planning algorithms, there is a plethora of hybrid motion planning techniques. Interestingly, they are being developed not only for individual vehicles but also for connected vehicles that make cooperative decisions for motion planning. To the best of our knowledge, there is a lack of a comprehensive review of the state-of-the-art (SOTA) \textbf{\textit{ hybrid motion planning methods for ADS}} that systematically analyses and compares existing recent studies in this field. Existing surveys have focused on research that uses either a non-hybrid for motion planning, e.g., data-driven methods~\cite{Teng2023, Aradi2022, schwarting2018planning}, or a collection of motion planning approaches without adequately emphasizing the importance of hybrid solutions~\cite{gonzalez2015review,claussmann2019review,katrakazas2015real,Long2023,paden2016survey, Long2023b}. This paper covers over $50$ hybrid methods for motion planning are explained, categorised, and compared. The categorisation logic is initially based on the components of the hybrid/combined method and subsequently on the challenges tackled by each hybrid approach. Furthermore, for every hybrid technique, particular attention is directed towards the interaction between the underlying TG and TDM processes. We hope this article will help researchers acquire a comprehensive understanding of how motion planning algorithms can be enhanced through the integration of diverse techniques.

The rest of the paper is organised in the following sections: After giving terminologies and definitions, Section~\ref{bckg} introduces the architecture of the motion planning module and related subsystems in AVs. In Section~\ref{classifying}, following the classification of the traditional methods, the hybrid techniques are reviewed, while their performances are discussed in Section~\ref{compare}.  %Key performance metrics and criteria alongside the discussion of 
%The performances of various hybrid and cooperative motion planning methods are discussed in Section~\ref{compare}. 
Section~\ref{conclusion} concludes the highlights of this survey. 
% \begin{landscape}
% \begin{table}[p]
% \renewcommand{\arraystretch}{1.3}
% \caption{Categorization, features, and comparison of various methods used in control module}
% \label{table1}
% \centering
% \includegraphics[width=1\linewidth]{table_1.pdf}
% \end{table}
% \end{landscape}

\section{Background on Motion Planning}
\label{bckg}
The modular system architecture is the commonly adopted model for designing and implementing automated driving systems (ADS), where several sub-systems are dedicated to different tasks of automated driving. The modular architecture breaks down the complex motion planning problem into tactical decision-making (TDM) and trajectory generation (TG). That simplifies the design of the motion planner, however, the interactions between TDM and TG must also be considered. A representative modular system architecture for the motion module of an ADS, including the interactions between global route planning, TDM and TG is illustrated in Fig.~\ref{fig:cont_layout}. The global route planner module is responsible for calculating the optimal route between the start and the end coordinates, through roads and pathways, using offline/online maps. The global route is usually made up of waypoints and does not provide any further detail, besides, it depends only on the current coordinate, endpoint, and feasible pathways. The global route planning is invoked only on-demand (see the input signal \textbf{A1} in Fig.~\ref{fig:cont_layout}), i.e., whenever following the designated route is not anymore feasible, e.g., due to an imminent traffic jam or roadworks. The other two main components, namely the TDM, or behaviour layer, and the TG interact with each other to provide the reference signals for the low-level control of the vehicle to energize its actuators (see the signals \textbf{D} and \textbf{O1} in Fig.\ref{fig:cont_layout}). 

Next, after defining key technical terms used in motion planning and I/O of the control system in an AV, the role of TDM and TG  modules are described in detail. 
 
\subsection{Terminology}
\label{sec:Terminology}
\begin{figure}[t]
\centering
\includegraphics[width=1\linewidth]{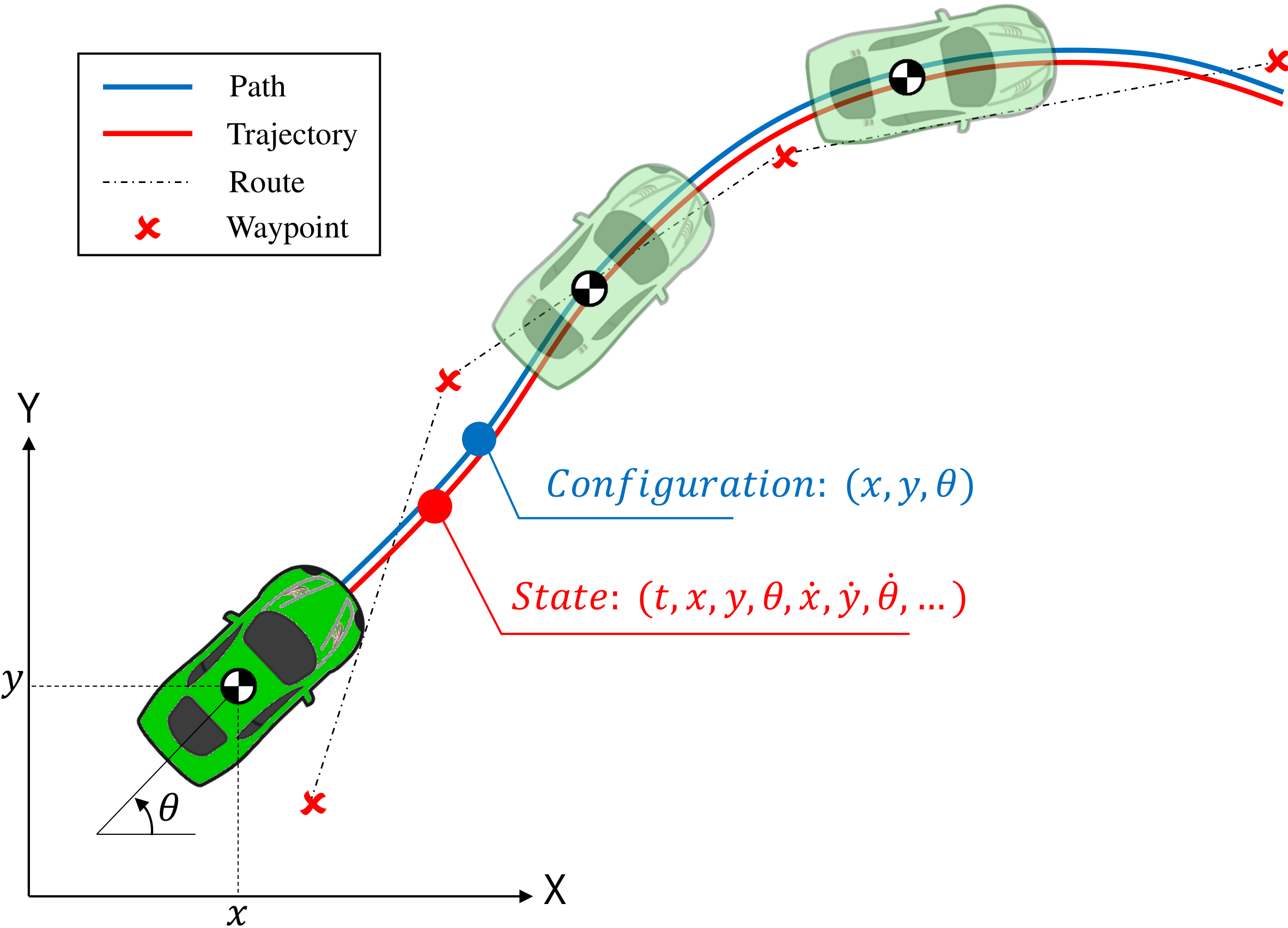}
\caption{Example illustration for the route, path, waypoints, and trajectory generation in a Cartesian global coordinate system. The state of the vehicle at time $t$ is defined in terms of its location $x, y$, orientation $\theta$ and their derivatives (e.g., acceleration, jerk, etc). The trajectory (red curve) and the path (blue curve) spatially coincide, but in this illustration, they are not drawn on top of each other for presentation clarity. The trajectory/path doesn't have to contain the waypoints, which only determine the high-level route planning.}
\label{fig:car}
\centering
\end{figure}

To have a common understanding of the terminology used throughout this survey, key technical terms for ADS are defined below. 
\begin{itemize}
\item \textbf{The configuration} of a vehicle in the 2D planar space includes its position and orientation, which uniquely determine where each point of the vehicle lies in space. The configuration could be represented in Cartesian (global or local) coordinates (Fig.~\ref{fig:car}), polar coordinates, Frenet frame, etc.     

\item \textbf{States} refer to the kinematics of the vehicle such as position, velocity, acceleration, etc. The vehicle's configuration can be a subset of its states.

\item \textbf{Configuration space} of a vehicle is divided into the following three subsets: Free space, collision space, and unknown space, according to the collision check of the vehicle in the physical space surrounding the vehicle.

\item \textbf{Path} is defined as a set of configurations from free (or unknown) space independent of any other variables like time.

\item \textbf{Trajectory} is defined as a set of states from the state space of the vehicle. The red curve in Fig.~\ref{fig:car} shows the trajectory and each point on this curve corresponds to a particular state (as a function of time). 

\item \textbf{Motion} refers to any change in the states of the vehicle. 

\item \textbf{Actions} are all possible control commands that could be applied to manipulate the motion of the vehicle.

\item \textbf{Maneuver} is defined by the characteristics or features of the vehicle's motion such as motorway merging, overtaking, turning, lane-change, etc.  

\item \textbf{EGO vehicle (EV)} refers to the controlled vehicle that the planning algorithms are designed for among all other surrounding vehicles and road participants. 

\item \textbf{Non-holonomic constraints} restrict the possible trajectories between two states because of the kinematic of the vehicle. In such systems, the trajectory leading to a state affects that state. In the case of AVs the 3D configuration (2D position and yaw orientation) is controlled by only two inputs (throttle/brake and steering angle). This occurs when the constraints cannot be integrated into the equation of motion~\cite{marsden1994on}.  

\item \textbf{Motion primitives} are a set of predefined/precomputed discrete trajectories that an EV can take from a given state.
\end{itemize}

%\subsection{Inputs, and Outputs}
Depending on the approach used to tackle the motion planning problem, the architecture of the control module and the required inputs to it could change accordingly. For instance, end-to-end autonomous driving integrates perception, prediction, planning, and control in a single unit. Therefore, the inputs to the motion planner in that case are raw sensor data, such as raw video captured by cameras, LiDAR point clouds, and radar readings. Whereas, in the modular approach, the raw data is first processed by the perception module to obtain a semantic representation of the environment, which is subsequently used by other modules, e.g., for prediction, risk assessment, control and motion planning. 

Even though the architecture of the modular approach is not unique, the key functions governing the structure of motion planning and control are the TDM (or behavioural layer) and the TG functions. These functions could operate either as two separate but interactive processes (comparable to a hierarchical planning design in DARPA Urban Challenge (DUC) for most of the participating vehicles \cite{buehler2009darpa}), or as a single integrated unit~\cite{gu2016automated,qian2019synchronous,esterle2018spatiotemporal}. In this survey, we cover the hybrid methods developed for one or both of these processes. Therefore, before categorizing the existing methods, the functionality of the two modules will be described in more detail. It is worth mentioning that the low-level control could either operate separately and follow the reference trajectory (the output of the TG module) using feedback control, or integrated within the planning section (similar to some MPC motion planning algorithms~\cite{rasekhipour2016potential}), whose output is directly employed to energize the actuators (signal~\textbf{O1} in Fig.~\ref{fig:cont_layout}). 

\subsection{Tactical Decision Making (TDM) function}
\begin{figure}[t]
\centering
\includegraphics[width=1\linewidth]{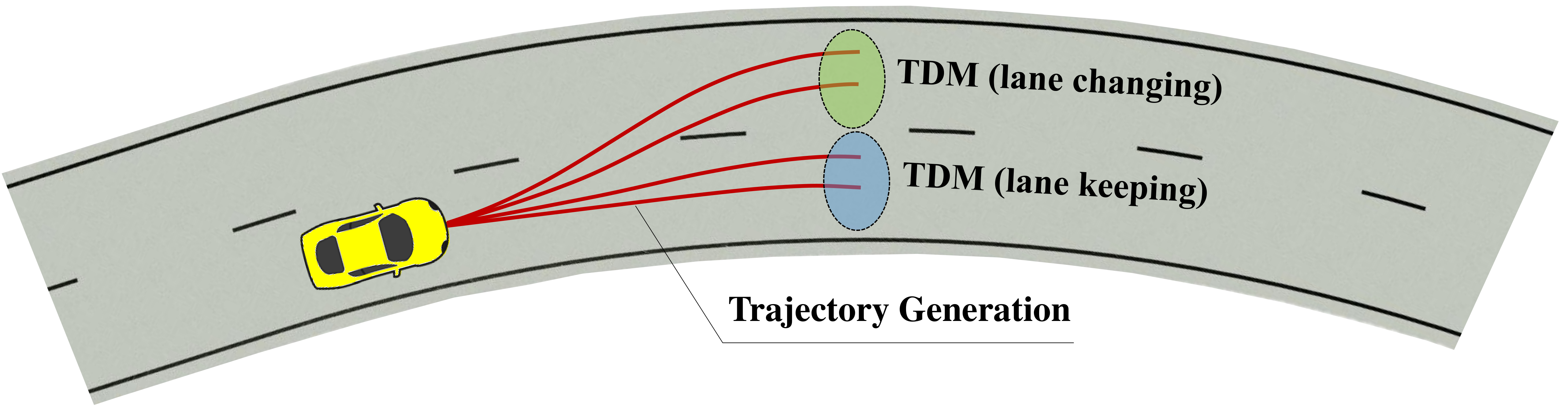}
\caption{Example illustration for manoeuvre-based TDM (lane-change or lane-keeping) and TG (two candidate trajectories per TDM output).}% for a simple driving scenario involving a two-lane road.}
\label{fig: tdm_tg}
\centering
\end{figure}

In principle, the TDM function is responsible for calculating the best behaviour for the vehicle according to the perceived driving context, e.g., deciding whether to do a lane-change or lane-keeping in the example illustrated in Fig.~\ref{fig: tdm_tg}.  Obtaining a comprehensive awareness of the surrounding area is a prerequisite step for properly reacting within a dynamic driving environment. Context-aware decision-making is the main difference between \textit{automation} and \textit{autonomy}, where the former concept refers to the ability to automatically control the behaviour of a system under certain conditions, whereas the latter also incorporates the ability of the system to properly react to unexpected changes within its environment. Traditional designs for the TDM layer are based on finite choices or decisions within a finite number of possible driving scenarios. Recent methods try to add autonomy in the design and address the behaviour of the vehicle in unexpected situations too, where the driving context cannot be classified under any of the predetermined states. Traditional and recent TDM methods for motion planning are further elaborated in Section~\ref{classifying}. Moreover, the TDM outputs could be used for other purposes too, such as activating intention signalling through visual displays to inform other road users about the manoeuvre intention of the EGO vehicle (signal~\textbf{O2} in Fig.~\ref{fig:cont_layout}) or optimizing the sensor coverage area by modifying the sensor's attention region of interest (signal~\textbf{O3} in Fig.~\ref{fig:cont_layout}) \cite{buehler2009darpa}.

\subsection{Trajectory Generation (TG) function}
As explained in Section~\ref{sec:Terminology}, the trajectory is generally defined as a ``path" that is followed by an object through space as a function of ``time". This is the reason why in some technical texts, the trajectory is also referred to as a  \textit{``spatiotemporal"} function~\cite{ziegler2009spatiotemporal,mcnaughton2011motion}. While the geometric representation for both the path and the trajectory in the spatial domain (drivable area) is the same (see Fig.~\ref{fig:car}), the trajectory includes additional kinematic (temporal) information (states) of the vehicle. In the case of AVs, the workspace or physical space (the road network) is usually planar (2D), and the configuration space is 3D, with two coordinates representing the position of the centre of gravity of the vehicle, and the third being the rotation about the normal axis. Different TG methods may require different representations of the physical space using, for instance, Voronoi tesselations, cost maps or state lattices~\cite{katrakazas2015real}. 
% Moreover, since there are only two control actions (steering and acceleration/deceleration), the kinematic motion is restricted to the non-holonomic constraints as well. 
The TG process is in charge of calculating the final signal to be tracked by the low-level control, see the signal~\textbf{D} in Fig~\ref{fig:cont_layout}, which applies the final action to the actuators of the AV. However, in (dynamic) model-based methods such as receding-horizon planning and control, the final feedback/feedforward control is integrated into the planning module and the output signal is directly fed into the actuators. In the following section, different categories for both traditional and hybrid approaches for TG are explained in detail. 

\section{Categorization of Existing Methods}
\label{classifying}
A great variety of algorithms and methods have been introduced in the last three decades for motion planning in AVs. These methods have been continuously evolving, which has resulted in researchers classifying them differently. This variety confuses, in some cases, the readers. In this section, we categorize, to the best of our knowledge, the existing methods, alongside the basis and logic behind their classification. Before focusing on hybrid approaches in Section~\ref{sec:hybrid}, it is essential to review the classical and learning-based methods for motion planning in Section~\ref{sec:TM} and Section~\ref{sec:AILBM} respectively, which are also the building  blocks of the hybrid algorithms. % developed more recently.
%will be explained in detail in the following subsections. 
% We also provide two tables to summarize the existing methods, along with related references. 

\subsection{Traditional Methods}
\label{sec:TM}
\begin{figure*}[t]
\centering
\includegraphics[width=1\linewidth]{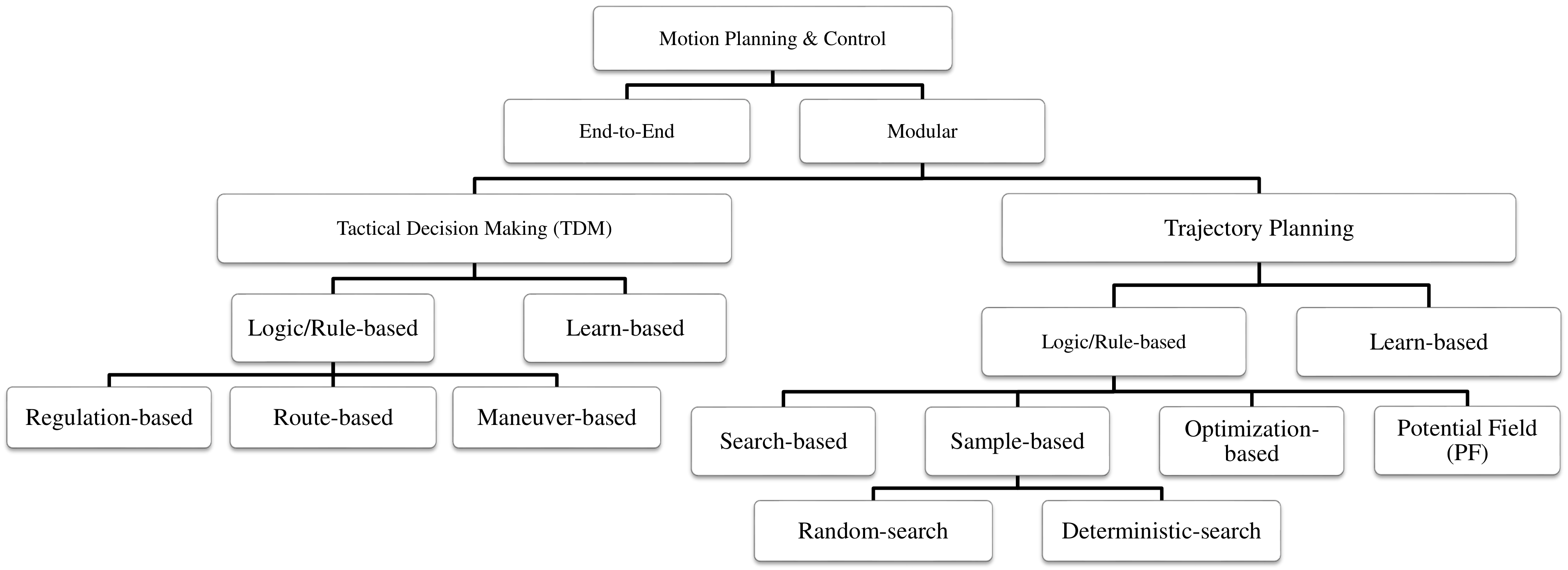}
\caption{Classification of the traditional methods (elements of the hybrid methods) used for motion planning and control system in AVs.}
\label{fig:chart1}
\centering
\end{figure*}

The traditional methods for TDM and TG processes are separately summarized and classified in the subsequent parts of this section along with their pros and cons, see  Fig.~\ref{fig:chart1} for the breakdown of traditional motion planning methods into different categories.
\begin{figure}[t]
\centering
\includegraphics[width=1\linewidth]{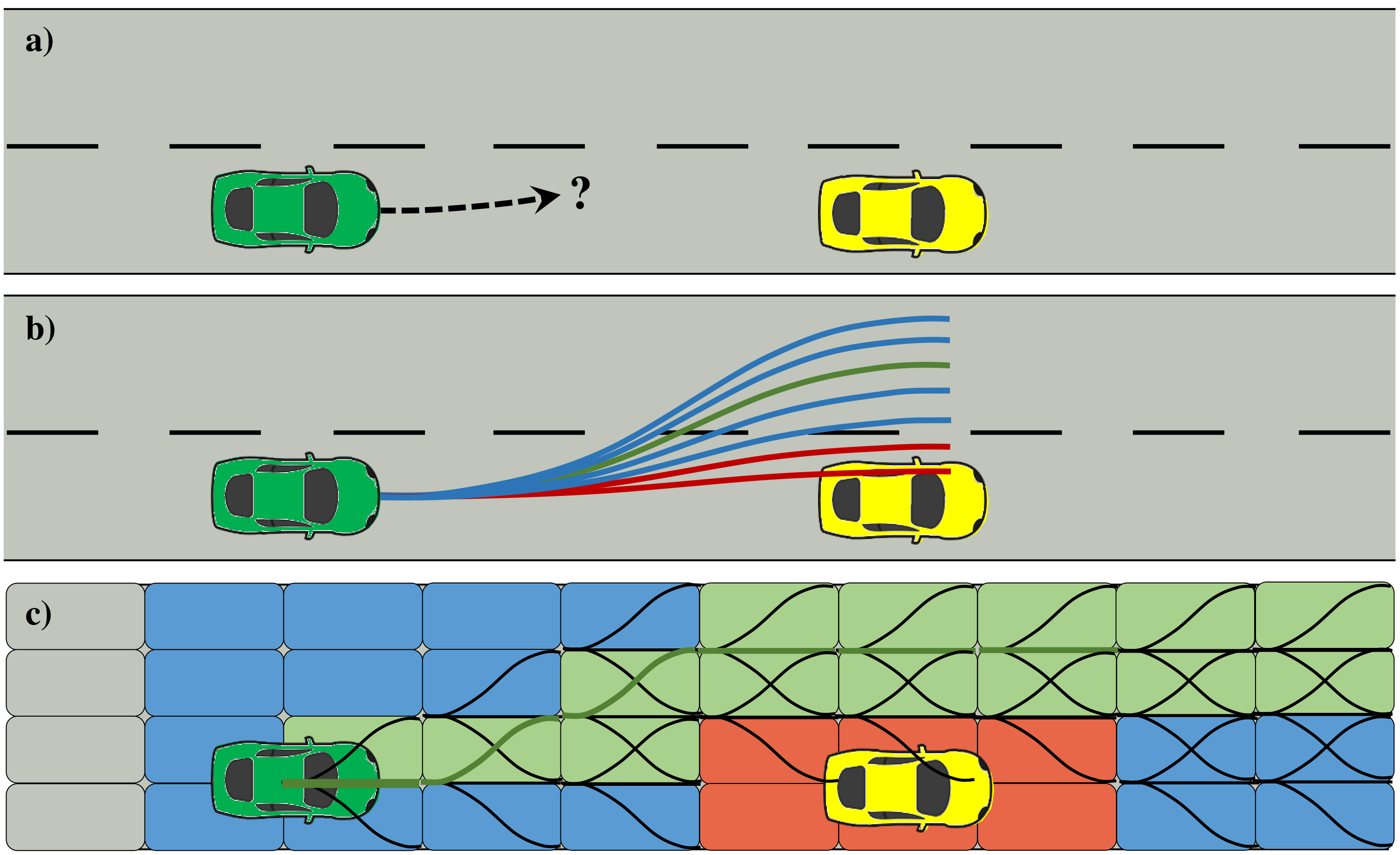}
\caption{Example illustrating the sampling-based (b) vs the search-based (c) method for a lane change scenario (a). The search space in this illustration is occupancy grids in (c) and trajectory samples in (b) that are generated via parametric curves. Colours show the candidate (blue), dismissed (red), and accepted (green) trajectories.} 
\label{fig: search_sample}
\centering
\end{figure}

\subsubsection{TDM Algorithms}
In the literature, one can find both learn-based and logic/rule-based methods for the TDM part of the motion planning module, with the rule-based methods being divided into the following categories~\cite{gu2017improved}:
\paragraph{Regulation-based Methods} They cover high-level rules and regulations, which are imposed by road authorities, such as stop/speed limit signs.
\paragraph{Route-based Methods} In this category, TDM is dictated by the higher-level global route planning module (signal A2 in Fig~\ref{fig:cont_layout}). For example,  selecting the exit in an intersection or a roundabout can be changed upon receiving extra information about imminent traffic jams, or roadworks. % or adverse weather conditions.
\paragraph{Manoeuvre-based Methods} They address WHEN and/or HOW to take an action. This type of tactical decisions are of utmost importance because they also interact with the TG function (signals B1 and B2 in Fig~\ref{fig:cont_layout}). %and subsequently the final control actions. %depends on them and both processes (TDM and TG) should be handled together.

Rule- and route-based TDM methods are handled directly by high-level modules such as online/offline driving regulations and global route planning systems, respectively, and they are referred to as automation-based decisions. These methods are usually designed and operate independently of the TG module. On the contrary, the manoeuvre-based TDM methods interact and affect the TG~\cite{gu2016automated}, and are referred to as autonomy-based decisions. There are several emerging challenges because of the interaction effects between TDM and TG, which become more difficult to address in the course of real-time control of the vehicle in complex environments. For instance, the output of the TG process may contradict the selected manoeuvre by the TDM, or alternatively, the TDM could make decisions without considering the feasibility/admissibility of the trajectories. Several studies have addressed this challenge by combining TDM and TG using topology-aware techniques~\cite{gu2016automated, qian2019synchronous} or by generating trajectories for each  manoeuvre group~\cite{esterle2018spatiotemporal}. 
%using a topology-aware technique for grouping together the candidate trajectories to preserve consistency, e.g, in obstacle avoidance~\cite{gu2016automated}, and lane changing~\cite{qian2019synchronous} scenarios, 
Hence, deciding about the manoeuvre without considering the connection between this process and the TG is not recommended. %and in general, the two modules should be treated interactively.
\\

\subsubsection{TG Algorithms}
TG algorithms are mostly based on local path planning. The first TG processes developed for AVs are similar to path planning methods, which is why some researchers stated that TG challenges could be handled by path planning techniques capable of dealing with differential constraints (accounting for the additional variable of time)~\cite{paden2016survey}. In a significant number of studies, the logic-based TG algorithms (Fig.~\ref{fig:chart1}) have been developed using one of the following four methods: \textit{Sample-based}, \textit{search-based}, \textit{optimization-based}, and \textit{potential field (PF)} approaches, which are further discussed and reviewed in the following paragraphs.

\paragraph{Sampling Methods} These methods were among the first algorithms developed for mobile robot motion planning experiences. Most of the modern references that implement a variation of this approach for AVs are inspired by DUC~\cite{kuwata2009real,mcnaughton2011motion,werling2012optimal,chu2012local,li2015real}. The sampling-based methods explore the environment via a sampling approach using a collision detection module to decide whether a sequence of samples constructs a valid configuration across the physical space. See Fig.~\ref{fig: search_sample}(b) for an example illustration. Sampling methods are computationally efficient, easy to implement, and could be further divided into \textit{random} and \textit{deterministic} sampling. \textbf{Random sampling} methods are usually incremental, such as the rapid-exploring random trees (RRTs), making them suitable for real-time applications. For example, in the RRT method, only a single point is sampled at a time (incremental sampling) and the algorithm has to decide whether to connect this point to the tree or not. However, for the models in which inertial properties are important, such as highly dynamic manoeuvres, the actual system in the real world does not usually follow the kinematic model. So, it would be intricate to implement algorithms like RRT without any modification. %Therefore, in the literature, some methods are proposed to overcome this drawback. 
In MIT's vehicle of \textit{Talos}, participated in the DUC, the RRT method was extended by augmenting a model-based closed-loop control to generate trajectories that are more appropriate for the dynamics of the vehicle \cite{kuwata2009real}. In \textbf{deterministic sampling} methods, the trajectory patterns are predefined according to the kinematic constraints of the vehicle or the road curvature. Different versions of state lattices \cite{ziegler2009spatiotemporal,mcnaughton2011motion} or candidate trajectories generated by differentiable polynomials (cubic, quintic, etc) are classified under this category \cite{li2015real,chu2012local}. For selecting the best trajectory, graph search methods are applied based on a cost function that evaluates the cost of each predefined piece of trajectory. Moreover, in order to avoid obstacles, the candidate trajectories colliding with obstacles are removed from the selection pool. The final trajectory is from a predefined discrete space, which means it is sub-optimal. Note that sub-optimality is the main drawback of sampling-based TG methods, because they are either incremental (random sampling), or use only a set of predefined patterns  (deterministic sampling).

% \begin{itemize}
% \item \textit{Random sampling:} Randomly build the search space to find the best trajectory/path, methods like RRT.

% \item \textit{Deterministic sampling:} Search grid patterns are predefined based, methods such as state lattice.
% \end{itemize}
\paragraph{Search Methods}\label{search} The search-based methods discretize the surrounding environment, e.g., the road network, by using a set of motion primitives (precomputed motions of the AV) and then applying graph-based search strategies such as Dijkstra\cite{gu2016runtime,bohren2009little} or A* family algorithms~\cite{kammel2009team,boroujeni2017flexible} to find the best path or trajectory according to a heuristic or objective of the motion planning.  See Fig.~\ref{fig: search_sample}(c) for an example illustration. In search-based methods, handling the kinematic constraints of the vehicle depends on the motion primitives used for constructing the search graph. Some modifications such as the hybrid state A* family of algorithms~\cite{dolgov2008practical} were proposed to take into account the nonholonomic kinematics in AVs. Unlike the sample-based methods, the search-based methods need to know in advance the topology of the road network. Despite that, the final trajectory might still be sub-optimal or non-smooth because of the discretized (graph-based) representation of continuous space. Increasing the size of the search graph or the resolution of the search grids improves the quality and the smoothness of the selected trajectory, however, the processing time also increases exponentially in that case and the computation load would be the main concern for real-time applications.  

\paragraph{Optimization Methods}
Optimization-based TG is designed to address the drawbacks of sample-based and search-based algorithms, i.e., sub-optimality and nonsmoothness, respectively. Besides, in deterministic sampling-based methods, the trajectory candidates are predefined and not conforming to a new environment. On the contrary, the optimization-based methods act in a continuous space and are more flexible with changing environments\cite{ziegler2014trajectory}. On the negative side, they are usually computationally expensive \cite{zhang2013dynamic,ziegler2014trajectory,gutjahr2016lateral,nilsson2016longitudinal,katrakazas2015real}, and they may cause unstable behaviour because of outcome oscillations from cycle to cycle of the computation~\cite{lim2019hybrid}.

\paragraph{Potential Field (PF) Methods}\label{potential} Like other traditional motion planning methods, the PF was introduced in the field of robotics. Khatib et al. proposed an obstacle avoidance motion planning method in which they assigned the repulsive and attractive PFs for the obstacles and the target, respectively and used them to navigate in the configuration space for manipulators\cite{khatib1978dynamic} and mobile robots\cite{khatib1986real}. The main advantage of the PF-based methods lies in their simplicity by providing an abstraction of complex environments with several obstacles, irregular geometries and boundaries making them desirable for motion planning algorithms. However, their main drawback, despite modified versions~\cite{bounini2017modified,lu2020hybrid}, is the existence of local minima traps, where repulsive and attractive forces cancel each other, thereby preventing the progress towards the goal. In ADS this situation could occur when another road user is located in between the EGO vehicle and the goal, which is often the case. Another problem with PF methods is the oscillation near obstacles and boundaries caused by gradient descent navigation. This issue has been addressed by adjusting the gradient~\cite{ren2006modified}, or modifying the control command by augmenting the proportional-derivative (PD) to the PF method (the input of the PD is the PF instead of the error from the reference trajectory). In this method, the PD controller coefficient could be used to tune the behaviour of the vehicle close to the obstacles\cite{galceran2015toward}. There are also other methods to compensate for the downsides of PF methods, which combine the PF with optimization-based methods like model predictive algorithms that will be later discussed in the section dedicated to the hybrid methods.  

\subsection{Artificial Intelligence and Learning-Based Methods}
\label{sec:AILBM}
Before delving into the review of the hybrid motion planning methods, we discuss, in this section, another category of motion planners (besides traditional), namely data-driven motion planners, which according to Fig~\ref{fig:chart1} can be either end-to-end or modular. As we will shortly see in the next section, data-driven modular motion planning can become one of the components used in hybrid methods. 

The popularity of data-driven methods has been following the recent breakthroughs in processing hardware and the proliferation of road traffic datasets under various driving scenarios~\cite{schwarting2018planning}. %another set of algorithms emerged for motion planning and control of AVs, which is generally known as learning-based methods. 
These methods can be broadly divided into two main classes, i.e., end-to-end or modular planning.  
The former refers to the methods in which the input features to the motion planner are sensory data such that collected from LiDAR, RADAR, GPS \cite{caltagirone2017lidar}, or camera~\cite{bojarski2016end,richter2017safe}, whereas, in the latter, perception and motion planning are done in separate modules. In the modular approach, the output of the perception module can serve as input for motion planning (see Fig.~\ref{fig:cont_layout}), providing, for instance, a semantic representation of the surrounding environment such as other road users represented by bounding boxes along with their velocities, as well as the positions of lane markings and the road geometry. In both modular and end-to-end methods, the output of motion planning could be a decision for triggering a maneuver~\cite{vallon2017machine, huy2013dynamic}, a state prediction of the surrounding environment~\cite{mozaffari2020deep}, a reference trajectory~\cite{lefevre2015learning}, or control commands like steering angle and throttle/brake activation signals~\cite{pomerleau1989alvinn, lu2019personalized}. 

Since the time-consuming learning phase is basically performed offline, learning-based methods are computationally efficient during real-time inference. Another desirable feature is that they are adaptive to diverse driving scenarios without significant change in the main structure given sufficient training data. The main challenge is that their performance depends on the quality and diversity of data provided during the training phase, which means that once the test situation deviates from the training conditions, the performance starts to degrade (or at least the level of reliability decreases). Furthermore, it is also difficult to debug the motion planning module in case of failure or improvement due to the well-known problem of explainability in data-driven techniques. 

 In order to overcome the challenges of the aforementioned classical algorithms, some other techniques have been developed, called \textit{``hybrid"} or \textit{``combination"} methods. In these methods, some of the drawbacks of the previously explained algorithms can be resolved by combining them in various hybrid frameworks, which can further improve the performance of motion planning and control for real-world applications.

\renewcommand{\arraystretch}{1.5}
\begin{table}
\centering
\caption{Comparative performance of the traditional TG methods against the five challenges described in Section~\ref{sec:hybrid}. Two check marks: best performance, and two crosses: worst performance in addressing the corresponding challenge.}
\label{table:traditional}
\begin{tabular}{lccccc} 
\toprule
\multicolumn{1}{c}{\multirow{2}{*}{\textbf{Traditional Methods}}} & \multicolumn{5}{c}{\textbf{Performance against challenges}}         \\ 
\cline{2-6}
\multicolumn{1}{c}{}                                              & \textbf{a}  & \textbf{b}  & \textbf{c}  & \textbf{d} & \textbf{e}   \\ 
\midrule
Searching Methods                                                & \cmark       & \cmark       & \xmark       & \xmark      & \xmark \xmark  \\ 
\hline
Sampling Methods                                                  & \cmark       & \cmark       & \xmark       & \xmark      & \xmark \xmark  \\ 
\hline
Optimization Methods                                              & \cmark \cmark & \cmark       & \xmark \xmark & \cmark      & \cmark        \\ 
\hline
Potential Field Methods                                             & \xmark       & \cmark \cmark & \cmark       & \xmark      & \cmark        \\
\hline
\end{tabular}
\end{table}

\subsection{Hybrid Methods} 
\label{sec:hybrid}
The majority of the traditional motion planning methods reviewed so far are not directly applicable to SAE level 3+ AVs, as they are not able to deal with all the motion planning objectives at once, such as low computational cost, global optimality, timely reactions to dynamic environments, adaptability to general driving scenarios, etc. In order to meet the requirements of highly ADS, recently published motion planning algorithms have been developed by combining various traditional methods. The resulting algorithms are known as hybrid motion planners  and they break down TDM and TG into simpler sub-problems or they try to address each motion planning objective separately by using properly-selected methods specially designed for that particular objective. This means that the drawbacks of a method can be compensated by another one, yielding a combined improvement to the overall performance. 

In order to classify and review the hybrid methods, we will first explore the challenges that a motion planning algorithm in an ADS needs to overcome. After that the hybrid methods will be categorized according to those challenges they are designed to address.  As illustrated in Fig.~\ref{fig:cont_layout}, it is assumed that the perception and prediction of the behaviour of other road users are provided to the motion planning system by other modules. In this review paper, we do not consider the impact of imperfect perception/prediction on motion planning which requires careful attention in its own right. % in the control module. 

\paragraph{\textbf{Vehicle's dynamics and feasibility challenges}} \label{ch.nonhol} The first challenge in designing a motion planning algorithm stems from the kinematic and dynamic constraints of the controlled vehicle. In the case of automated driving, these constraints include the nonholonomic kinematics of the front wheel steering system and the control effort saturation of both magnitude and rate of change in throttling/braking and steering actuators. Even in an open, flat drivable area without any other participants, the motion planning system still has to cope with this challenge. Ignoring the vehicle's kinematic constraints may lead to unfeasible reference trajectories, followed by more errors in the low-level controller.

% \begin{table}[t]
% \centering
% \caption{Performance of the traditional motion planning methods against challenges described in sections~\ref{ch.nonhol} to~\ref{ch.uncertainty}}. 
% \label{table_traditional}
% \includegraphics[width=1\linewidth]{table_tradition.pdf}
% \centering
% \end{table}

\paragraph{\textbf{Driving context challenges}} This group is related to the environmental constraints enforced either by the structure of space and the driving scenario, or by other road participants affecting the driving context of the EGO vehicle (EV). The driving environment could be divided into structured (predefined routes such as roads and highways) and unstructured spaces (parking lots, and off-road). The structured environments can be further divided into the major classes of urban and highway driving, where urban scenarios may include intersection and roundabout crossings, traffic jams, etc, and highway scenarios may refer to overtaking, lane changing, lane keeping, highway merging or exiting. The context-based driving challenges play a crucial role in determining which motion planning method is suitable for a particular situation. One way to overcome this challenge is to develop general-purpose motion planning methods that can operate in a wide range of driving situations including unexpected events. However, achieving this goal is not trouble-free as highlighted by existing methods in the literature, which usually assume that the driving scenario is made of a set of finite pre-determined manoeuvres and is subsequently provided as input for the TG and TDM processes. The authors  in~\cite{claussmann2019review} infer that the most promising algorithms for motion planning in highway driving are the parametric/semi-parametric curve planning methods (sampling-based TG). As concluded over there, other logic-based planning methods are not appropriate to be used standalone for TG in highway scenarios and using a \textit{hybrid} algorithm is recommended.
%Some of the algorithms in the literature are applicable to more than one scenario. The highway scenarios where the average speed of vehicles and the change in road curvature are defined in a determined range (speed over 60 km/h, and unidirectional road traffic flow) have been extensively elaborated in another review paper~\cite{claussmann2019review}.
\begin{figure}[t]
\centering
\includegraphics[width=0.9\linewidth]{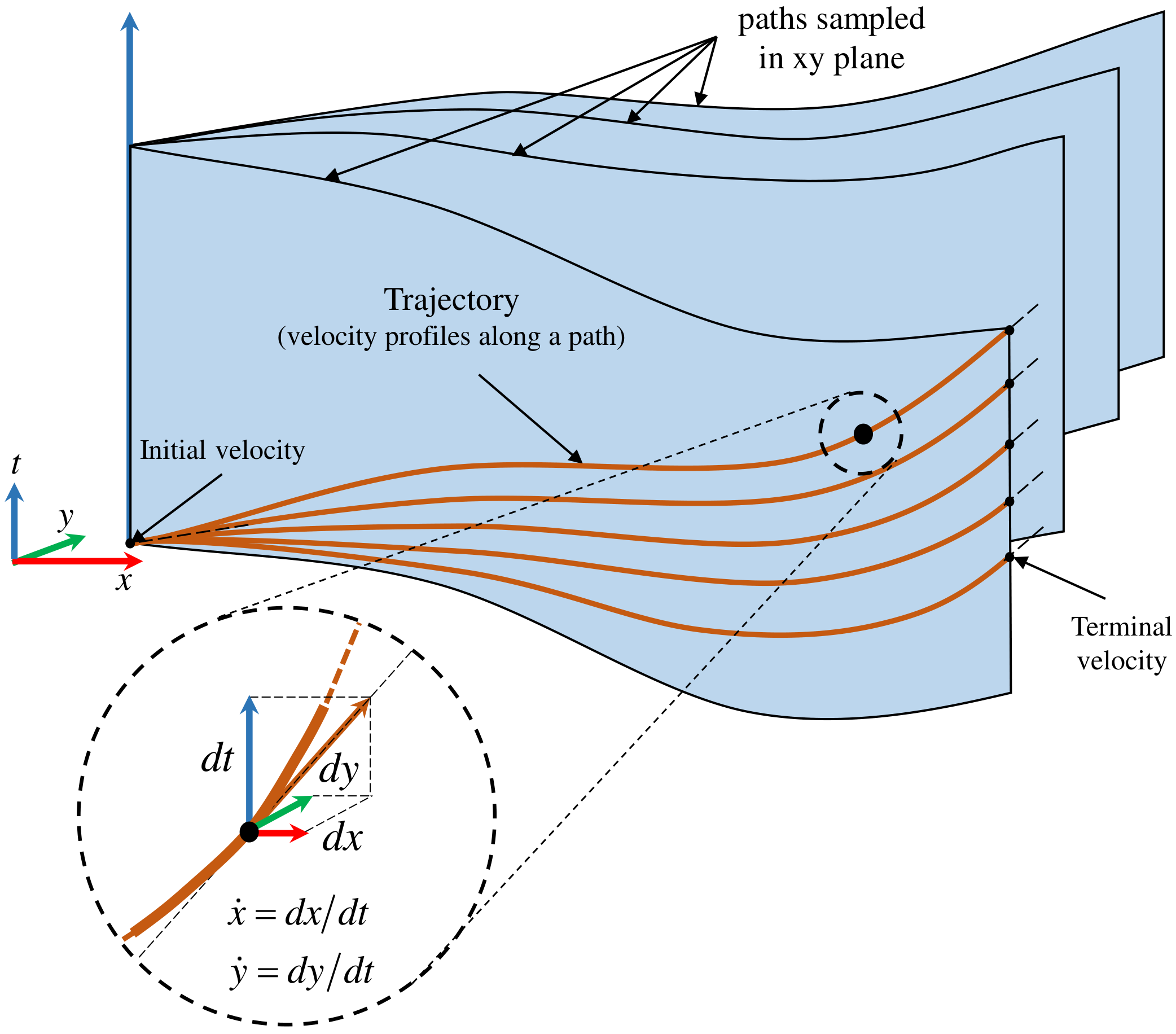}
\caption{Example illustration for the decomposed geometric and kinematic motion planning. Different methods could be used for generating the  geometric paths (blue surfaces) and the associated velocity profiles (brown curves). Given the 2D selected path (the front blue surface), different trajectories (brown curves), with the same initial/terminal velocities and different journey times are illustrated. }
\label{fig: decomposed}
\centering
\end{figure}

\paragraph{\textbf{Real-time implementation challenges}}\label{ch.realtime} The next source of challenges arises from the need for planning and implementing the motion of the vehicle in such a way that the EV could take actions and react to the environment as fast as possible, i.e., the computations in the TG and TDM modules should be performed in real-time. Given that the update rate of contemporary sensors is in the order of $100$ ms, a real-time algorithm should update at least once in a time period of $100$ ms (operate at a rate of $10$ Hz or higher). There is absolutely no value for a method that can produce the safest, most reliable and traffic-efficient outputs applicable in a wide range of driving contexts, but its processing time is slower than the state alteration rate of the EV, or the dynamics of the surrounding environment. Accordingly, the update rate of the motion planning module is one of the parameters determining the maximum travel speed of an AV. 

\paragraph{\textbf{Safety-based challenges}}\label{ch.safety} One of the most important metrics for evaluating motion planning methods, is the level of safety provided, which is explicitly considered in the design of the algorithm. Safety can be measured using several well-established key performance indicators such as the (modified) time-to-collision (TTC), the time and distance headways to surrounding vehicles, and in some scenarios the post-encroachment time within a conflict zone. At the same time, it is desirable that improving safety in an ADS should not compromise the stability of the controlled system, the passenger comfort (ride quality), and should not increase the number of unnecessary takeover requests to the safety driver (fail-safe control of the vehicle). For some methods in the literature, the safety metrics are directly incorporated in the mathematical formulation of motion planning problem, while others investigate the safety criteria by using simulations or experimental evaluations. 

\paragraph{\textbf{Uncertainty-based challenges}}\label{ch.uncertainty} The last challenge for developing a motion planning algorithm is how to handle the uncertainties due to any kind of imperfections in the input data provided by other modules, such as perception, prediction, localization, etc. Gaining robustness against various system faults can also further enhance the ride quality and safety levels offered by the motion planning algorithm. 

The performance of the traditional motion planning approaches evaluated against the above-mentioned challenges is summarized in Table~\ref{table:traditional}. By inspecting this table, it becomes clear there is no single method that addresses all challenges, e.g., optimisation-based approaches perform well in terms of feasibility, adaptability to the driving environment, safety and resilience to uncertainty, but their performance  may significantly compromise in a real-time ADS implementation. This issue has recently sparked a lot of research efforts in the so-called hybrid techniques, which combine various methods together with the aim to improve the performance over the classical approaches.  Next, the SOTA hybrid methods for motion planning will be explained, categorized and assessed against the same list of challenges. The results of our review study are also presented in Table~\ref{table:hybrid}.
\\

\subsubsection{Decomposed Geometry and Kinematic Planning}\label{hybrid_sampling} The most common way to simplify the TG problem, is to separate the geometric path from the kinematic properties of the trajectory, such as velocity, longitudinal and lateral acceleration. In this way, the complicated spatiotemporal TG problem breaks down into two simpler problems, i.e., path planning and velocity profile planning, which could be treated, in a hierarchical design model, see~Fig.~\ref{fig: decomposed} for an example illustration. For instance, Li et al. in~\cite{li2015real} generate candidate trajectories employing the cubic B-spline to initially refine the input reference path from the high-level route planner, and after that, to address the safety and comfort concerns, a velocity profile is generated by explicitly considering kinematic constraints such as longitudinal/lateral, speed/acceleration limits and traffic regulations. This approach leads to a closed-form solution for velocity profile planning, which can significantly reduce the real-time process capability of the proposed algorithm (the update rate achieved is approximately $70$~ms). The same concept is used for motion control of the VisLab AV in~\cite{broggi2012autonomous} by first generating the circular geometrical path and then calculating the speed profile based on the minimum and maximum curvature of the path to finally select the best trajectory in terms of comfort (lateral acceleration) and EV's input control limits (maximum steering angle). This approach is similar to multi-level motion planning used for nonholonomic indoor mobile robots, where the curvature and velocity features of the trajectory are improved in a sequential manner~\cite{simba2016real}. In another study~\cite{medina2023ia}, a data-driven method is used to generate waypoints, which are used as a reference to sample the geometric path based on Bezier curves, and after that, the velocity profiles are calculated for each path. This method has been used for challenging urban scenarios such as roundabouts, intersections, and T-junctions with  $3$~s and $6$~s time horizons with a minimum update rate of $4$~Hz. 

The decomposed geometry and kinematic planning approach makes it possible to use different methods for each part. For a highly constrained driving context, Zhang et al~\cite{zhang2018hybrid} decompose the TG  problem into a multi-layer planner including two steps: A geometry-based path generation and an optimization-based speed planner phase (Fig.~\ref{fig: decomposed}). The authors successfully reduce the computational load, while according to the simulation results also improve the motion planning performance in terms of both smoothness and magnitude of the curvature/heading rate, as compared to the hybrid A* algorithm used in~\cite{kato2015open} and~\cite{dolgov2008practical}. Similarly, Artunedo et al.\cite{artunedo2019real, artunedo2021jerk} develop a real-time motion planning for ADS in constrained driving environments like urban scenarios, by first using quintic Bézier curves to generate the path and then planning for the longitudinal speed by taking into account the limitations imposed by the curvature of the path, speed rules, acceleration limit, etc. The low complexity of this approach makes it suitable for developing collision avoidance emergency (fail-safe) trajectories like that studied  in~\cite{lee2019collision}. Over there the acceleration profile is planned both for the longitudinal/lateral directions based on the safest geometrical path, which is selected by assessing the collision risk along each candidate path. Another advantage of this approach is the flexibility and the convex properties of the formulated optimization problem used for the motion planning design as illustrated in~\cite{zhang2018toward}. The focus of this study is to optimize the speed profile along a continuous geometrical curve provided by another process in a hierarchical motion planning system. This procedure ensures that the optimization process will take place in a limited time span and satisfy the real-time implementation challenges. Finally, Lim et al.~\cite{lim2019hybrid} propose a hybrid TDM and TG method, which takes advantage of both sampling and optimization techniques. Firstly, the authors find the best on-road scenario by sampling for the lateral movement of the vehicle and secondly, they use a linear model predictive control (MPC) design for optimizing the longitudinal speed profile constrained to the previously-calculated safety corridor. 

The methods used to develop the aforementioned algorithms are a hybrid of the \textit{sampling}-based method for generating the geometrical path (see the blue surfaces in Fig.~\ref{fig: decomposed}) and the \textit{optimization}-based method for calculating the best speed profile (see the brown velocity profiles on the path surface in Fig.~\ref{fig: decomposed}). In other words, decomposition helps improve the sampling of the candidate trajectory by doing first intelligent sampling in the driveable area~\cite{gu2013focused}, and then planning a speed profile via the solution of a simplified optimization problem, which leads to a more computationally efficient outcome.
\\

\subsubsection{Potential Field-Based Hybrid Methods}\label{hybrid_pf}
The artificial potential field (PF) method is an appropriate tool for modelling various elements affecting the TDM and TG processes of the AV. However, there are also challenges associated with this method, as reviewed in Section~\ref{potential}, which can be alleviated by combining PFs with other motion planning algorithms.

\begin{figure}[t]
\centering
\includegraphics[width=1.0\linewidth]{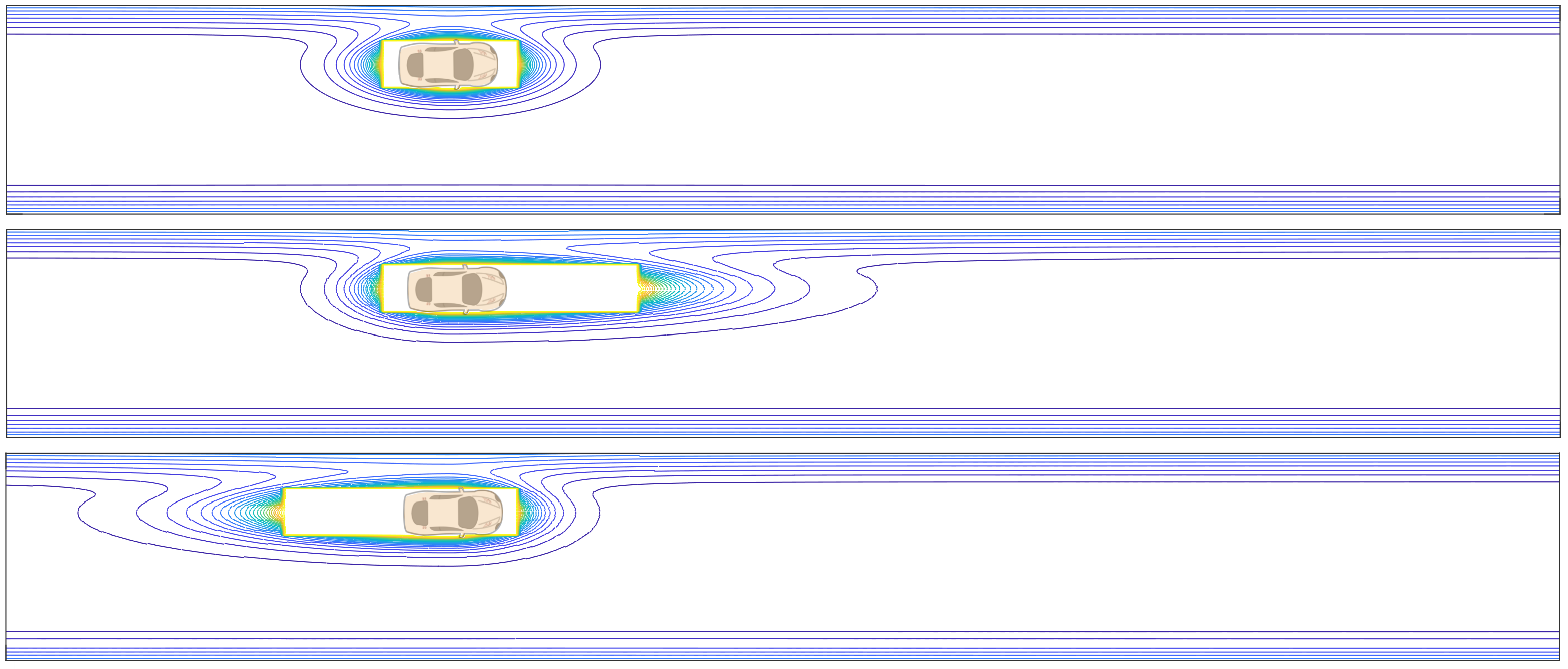}
\caption{A repulsive potential field generated for an obstacle (another vehicle) moving with a speed (equal, faster, and slower) than the EV. The PF is used as part of the cost function in the MPC motion planner in~\cite{rasekhipour2016potential}. The colour-coding of the contours, blue to red, is associated with low and high PF magnitude, respectively.}.
\label{fig: pf}
\centering
\end{figure}

Most hybrid motion planners based on the PFs are structured to simplify the optimization process. For instance, instead of adding inequality constraints to describe complex driving scenarios in optimization-based techniques, which prohibitively increase the computational complexity of the formulated optimization problem, a PF could be used as a penalising factor in the cost function. Several studies have concentrated on creating a hybrid motion planning approach based on PFs, wherein the PF is designed to be convex to streamline the optimization process~\cite{qin2023hierarchical, rasekhipour2016potential}.  Rasekhpour et al. in~\cite{rasekhipour2016potential} define a quadratic form repulsive PF (non-crossable/crossable) as a function of the relative distance and velocity between the EV and other participants (Fig.~\ref{fig: pf}), in addition to the lane markings (see also~\cite{liu2017path,ji2016path}), and road boundaries. The resulting PF is used in the MPC cost function for TG, and tested in different scenarios including merging, lane changing, and overtaking. Dixit et al. in~\cite{dixit2019trajectory} use a PF-like function to define a safe zone as a reference for MPC controllers in highway overtaking manoeuvres. By using so, they compensate for the drawbacks of both PF and optimization-based methods and guarantee the feasibility of the planned trajectory while modelling various road elements such as lanes, borders, surrounding cars, and their kinematic information. A similar framework is developed in~\cite{li2021combined} to add human-like driving habits (aggressiveness/cautiousness) to the motion planning module of the AV, while preserving the optimal trajectory thanks to the MPC part of the algorithm. The hybrid approach combining MPC with PFs has been also used to mitigate the severity of crashes on occasions where the accident is inevitable~\cite{wang2019crash}. Hang et al. in~\cite{hang2020human} combined MPC and PFs with a game-theoretic framework to model human-like decision-making behaviour. This aspect could be important for mixed traffic conditions where AVs coexist with human-driven vehicles.

The representation of the environment using PFs have also improved the performance for other TG methods. Huang et al. in~\cite{huang2019motion} introduce a novel approach for motion planning, by proposing a conductive state lattice grid structure (considering also the nonholonomic constraints of the vehicle, similar to the search-based methods discussed in section~\ref{search}), in which a resistance is assigned to each edge according to the value of the PF in the middle of that edge. Afterwards, by adding a voltage source between the current location of the AV and the local target point, the path is constructed by following the maximum electrical current route from the starting to the endpoint. Given the path, the velocity is calculated based on the decomposed geometry and kinematic planning discussed in the previous section. In another study, Park et al. in~\cite{park2021online} designed a hybrid algorithm for trajectory planning along curved roads with multiple obstacles (urban driving situation). They used the PF of the obstacles to replace the cost model used in the hybrid A* algorithm and improve its real-time performance.

In summary, the main idea behind PF-based hybrid methods for motion planning is to use PFs alongside optimization- or search-based methods to describe challenging environments such as urban driving scenarios with curved roads, (non-convex) boundaries and dense traffic at low complexity. Simply using a traditional optimisation- or search-based method with a high-resolution dense grid would incur a much higher computational effort. 
\\

\renewcommand{\arraystretch}{1.5}

\begin{table*}
\centering
\caption{Hybrid motion planning methods: {\textbf{1}} $\equiv$ Decomposed geometry and kinematic planning, {\textbf{2}} $\equiv$ PF-based hybrid methods, {\textbf{3}} $\equiv$ Optimization-based hybrid methods, {\textbf{4}} $\equiv$ Combination of logical and learning-based methods, {\textbf{5}} $\equiv$ Hybrid cooperative planning. The challenges (a to e) against which the performance of the different methods is assessed are: a) Vehicle's dynamics and feasibility, b) driving context, c) real-time implementation, d) safety-based and e) uncertainty-based.}
\label{table:hybrid}
\begin{tabular}{>{\centering}p{.5cm}p{1.6cm}ccccccccccccp{6.5cm}} 
% {>{\centering}p{0.07\textwidth}p{0.19\textwidth}>{\centering}p{0.07\textwidth}>{\centering\arraybackslash}p{0.05\textwidth}}
\toprule
\multirow{2}{*}{\textbf{Method}} & \multicolumn{1}{c}{\multirow{2}{*}{\textbf{Ref}}} & \multicolumn{7}{c}{\textbf{Classical Method Elements}}                                                                                                                                                                                    & \multicolumn{5}{c}{\textbf{Challenges}}                        & \multicolumn{1}{c}{\multirow{2}{*}{\textbf{Notes}}}                                                                        \\ 
\cline{3-13}
                                        & \multicolumn{1}{c}{}                              & \begin{sideways}Sampling\end{sideways} & \begin{sideways}Searching\end{sideways} & \begin{sideways}Optim.\end{sideways} & \begin{sideways}PF\end{sideways} & \begin{sideways}Fuzzy\end{sideways} & 
                                        \begin{sideways}Other Logics\end{sideways} &
                                        \begin{sideways}Learning\end{sideways} & \textbf{a} & \textbf{b} & \textbf{c} & \textbf{d} & \textbf{e} & \multicolumn{1}{c}{}                                                                                                       \\ 
\midrule
\textbf{1}                         &\cite{li2015real} \cite{lim2019hybrid} \cite{simba2016real} \cite{zhang2018hybrid} \cite{artunedo2019real} \cite{artunedo2021jerk} \cite{lee2019collision} \cite{zhang2018toward} \cite{gu2013focused}  \cite{medina2023ia}                  &\cmark                                     &                                         &\cmark                                   &                                  &                                     &         &                                 &\cmark         &            &\cmark         &\cmark         &            & the geometric path is designed using parametric curves                                                                     \\ 
\hline
\multirow{3}{*}{\textbf{2}}             & \cite{broggi2012autonomous} \cite{rasekhipour2016potential}\cite{liu2017path}\cite{ji2016path}\cite{dixit2019trajectory}\cite{li2021combined} \cite{hang2020human} \cite{qin2023hierarchical}                                   &                                        &                                         &\cmark                                   &\cmark                               &                                     &        &                                   &            &\cmark         &\cmark         &\cmark         &            & PF is the key to model the surrounding environment                                                                         \\ 
\cline{2-15}
                                        & \cite{wang2019crash}                                                &                                        &                                         &\cmark                                   &\cmark                               &                  &                     &                                        &            &            &            &\cmark         &            & developed to mitigate crash severity                                                                                       \\ 
\cline{2-15}
                                        & \cite{park2021online}                                                &                                        &\cmark                                      &                                      &\cmark                               &                                     &    &                                       &            &            &\cmark         &            &            & for urban driving situation                                                                                               \\ 
\hline
\multirow{3}{*}{\textbf{3}}             & \cite{lattarulo2021hybrid} \cite{hidalgo2019hybrid}                                             &\cmark                                     &                                         &\cmark                                   &                                  &                                     &                          &                &            &            &\cmark         &\cmark         &            & \multirow{2}{*}{the parametric curve could be changed by MPC optimization}                                                 \\ 
\cline{2-13}
                                        & \cite{gu2016road} \cite{zhang2021unified}                                               &\cmark                                     &                                         &\cmark                                   &                                  &                                     &                   &                       &\cmark         &\cmark         &\cmark         &            &            &                                                                                                                            \\ 
\cline{2-15}
                                        & \cite{ding2019safe} \cite{xin2021enable}  \cite{chen2022rrt}                                           &                                        &\cmark                                      &\cmark                                   &                                  &                                     &                       &                   &            &\cmark         &\cmark         &            &            & the search space is reduced by finding a 2D spatiotemporal corridor                                                           \\ 
\hline
\multirow{5}{*}{\textbf{4}}             & \cite{li2018humanlike} \cite{wang2021imitation} \cite{ma2020deepgoal}                                          &                                        &                                         &                                      &\cmark                               &                                     &                 &    \cmark                  &            &\cmark         &\cmark         &            &            & PF concept used either to ensure the control commands are safe or  reduce the complexity of learning based algorithms  \\ 
\cline{2-15}
                                        & \cite{wang2019quadratic}                                                &                                        &                                         &                                      &                                  &                                     &              & \cmark                        &            &\cmark         &\cmark         &\cmark         &            & combines RL with traditional PID controller; PF-like term in reward function is used to tune the PID coefficients                                                                                       \\ 
\cline{2-15}
                                        & \cite{hegedus2019motion} \cite{tuatulea2020design} \cite{sun2018fast} \cite{huang2023differentiable}  \cite{liu2023occupancy}  \cite{yang2022hybrid}                                     &                                        &                                         &\cmark                                   &                                  &                                     &       &   \cmark                              &            &\cmark         &\cmark         &            &            & computationally inefficient algorithm used to facilitate training process of learning-based algorithms                              \\ 
\cline{2-15}
                                        & \cite{wang2020learning}  \cite{zhang2022integrating} \cite{wen2023tofg}                                               &\cmark                                     &                                         &                                      &                                  &                                     &  
  &           \cmark                      &            &\cmark         &\cmark         &            &            &    The sampling is guided via learning based methods                                                                                                                        \\ 
\cline{2-15}
                                        & \cite{chen2020conditional}                                                &                                        &                                         &                                      &                                  &\cmark                                  &                 &   \cmark                   &\cmark         &\cmark         &            &            &            & the output of learning-based method are enhanced by modelling the relation between them using fuzzy logic                       \\ 
\cline{2-15}
                                        & \cite{sormoli2023novel} \cite{sun2023get}                                               &                                        &                                         &    \cmark                                   &                                  &                                  &          \cmark        &                     &\cmark         &\cmark         &            &            &            & the fluid flow dynamics and dynamic graph used for modelling interaction between actors                       \\                                         
\hline
\multirow{3}{*}{\textbf{5}}             & \cite{eilbrecht2018optimization} \cite{viana2021comparison}                                             &\cmark                                     &                                         &\cmark                                   &                                  &                                     &               &                           &            &\cmark         &            &\cmark         &\cmark         & individual cooperative motion planning                                                                                     \\ 
\cline{2-15}
                                        & \cite{hidalgo2021platoon} \cite{huang2018path}                                             &\cmark                                     &                                         &\cmark                                   &\cmark                               &                                     &               &                           &            &\cmark         &            &\cmark         &\cmark         & batch planning (platooning)                                                                                                \\ 
\cline{2-15}
                                        & \cite{huang2018path} \cite{ma2020novel}                                             &                                        &                                         &                                      &                                  &                                     &              &                            &            &\cmark         &            &\cmark         &\cmark         & combined TG and TDM using hybrid automata                                                                  \\
\hline
\end{tabular}
\end{table*}

\subsubsection{Optimization-Based Hybrid Methods} Some hybrid motion planning methods combining optimization-based approaches with PFs and sampling techniques have already been discussed in Section~\ref{hybrid_pf} and Section~\ref{hybrid_sampling}, respectively. However, there are more examples of hybrid approaches, in which the performance of an optimization-based technique was improved by combining it with another motion planning algorithm. These examples are reviewed next. %that do not fit into the previous categories. 

The main challenge of optimization-based methods is their high computational overhead, which can make them unsuitable for real-time implementations. One way to mitigate this challenge is to incorporate another method into the optimization-based framework. Lattarulo et al. in~\cite{lattarulo2021hybrid} proposed a hybrid motion planning approach consisting of two main steps: First, a smooth nominal trajectory is calculated, and next the trajectory/manoeuvre is optimized taking into account the constraints imposed by obstacles, and road conditions using an MPC architecture with the decoupled point-mass dynamical model. In a similar approach, Wonteak et al. in~\cite{lim2018hierarchical} developed a hierarchical motion planning algorithm (behaviour and trajectory) that reaps the advantages of both optimization and sampling methods. Specifically, the sampling algorithm is responsible for determining a high-level rough behaviour-based trajectory, and then a low-level optimization-based trajectory is generated accounting for the dynamic constraints introduced by the vehicle and the environment. Essentially, sampling is first used for TDM and then optimization is adopted for TG. In this framework, the high performance of the optimization method is mostly preserved, while the overall implementation complexity remains low. Hidalgo et al. in~\cite{hidalgo2019hybrid} improved the performance of motion planning in roundabout merging scenarios by combining the parametric curve (Bézier curve) for path planning with MPC for longitudinal and lateral control. They successfully reduced the overall computational cost while maintaining the high performance attained by MPC. Gu et al. in~\cite{gu2016road} developed a multi-layer framework for TG. In the first step, the authors optimized a traffic-free rough trajectory (curve and velocity profile), and in the next two steps, they generated the final smooth trajectory, incorporating the behaviour of other traffic participants as constraints. Ding et al. in~\cite{ding2019safe} and Zhang et al.~\cite{zhang2021unified} simplified the optimization problem by defining the drivable environment as a function of time using safe spatiotemporal corridors (SSC). Therefore, the computational complexity of TG can be reduced as the search space is confined  within the SSC. The performance of the proposed hybrid method was experimentally evaluated and successfully compared with other methods developed in~\cite{werling2012optimal}, showing promising results. In a similar manner, Xin et al. in~\cite{xin2021enable} first find a reference non-smooth trajectory in a 3D spatiotemporal map using a search-based algorithm (see Section~\ref{search}), and in the next phase, MPC is employed to smooth out the selected trajectory. Finally, in~\cite{receveur2020autonomous}, a genetic algorithm is combined with PF-based methods to enable optimized motion planning in real-time. The PF part, due to its simplicity and accuracy in capturing the changes in dynamic environments, enables reactive planning, while the genetic algorithm part ensures the optimality of the calculated trajectory. Furthermore, conventional path planning techniques like Rapidly Exploring Random Tree (RRT), which may exhibit suboptimal performance on their own, can be employed in conjunction with computationally intensive optimization methods~\cite{chen2022rrt}. The idea behind this approach is to prune the solution space before searching for the optimal solution.

To summarize, the main idea behind optimization-based hybrid methods is to facilitate the time-consuming optimization process by carefully reducing the size of the solution space. This is usually achieved by pruning some of the feasible driving behaviours using another (traditional) motion planning method before an optimisation-based technique is called to generate the optimal trajectory within the search space. 
\\

\begin{table*}[t]
\centering
\caption{Categorization of the hybrid motion planning methods based on the TDM and TG design approach and their interaction.}
\label{table:interaction}
\begin{tabular}{>{\centering}p{1.8cm}>{\centering}p{2.2cm}p{3.4cm}p{6cm}}
\toprule
\multicolumn{2}{c}{\textbf{Category}} & \multicolumn{1}{c}{\textbf{Reference}} & \multicolumn{1}{c}{\textbf{Definition}} \\ \midrule
\multirow{2}{*}{Separate} & TG & { \cite{li2015real, broggi2012autonomous, simba2016real, zhang2018hybrid, artunedo2021jerk, lee2019collision, zhang2018toward, gu2013focused, rasekhipour2016potential, ji2016path, dixit2019trajectory,li2021combined, wang2019crash, park2021online,lattarulo2021hybrid, tuatulea2020design} } & Continuous TG given a reference route or terminal state of the EV, regardless of long-term behaviour. \\ \cline{2-4} 
 & TDM & { \cite{hang2020human, receveur2020autonomous, li2018humanlike, ma2020deepgoal, wang2020learning} } & High-level TDM (discrete) determining behaviour, independent of the short-term trajectory of EV.\\ \hline
\multirow{2}{*}{Interactive} & Implicit & { \cite{liu2017path, gu2016road, wang2019quadratic, hidalgo2019hybrid, ding2019safe, chen2020conditional, zhang2021unified, huang2023differentiable, sun2023get, liu2023occupancy, wen2023tofg, chen2022rrt, qin2023hierarchical, sormoli2023novel} } & Simultaneous TDM and TG design without distinguishable in-between interactions. \\ \cline{2-4} 
 & Explicit & { \cite{gu2016automated, lim2019hybrid, xin2021enable, wang2021imitation, sun2018fast, ma2020novel, artunedo2019real, hegedus2019motion, huang2018path, medina2023ia, zhang2022integrating, yang2022hybrid} } &  Simultaneous TDM and TG design with distinguishable in-between interactions. \\ \hline
\end{tabular}
\end{table*}

\subsubsection{Combination of Logical and Learning-Based Methods} Recently, Artificial Intelligence (AI)-based methods are being used along with other well-known motion planning algorithms in a hybrid framework to improve the overall performance. In this section, we review various types of logic-learning hybrid motion planners in the following paragraphs. 
%The subsequent paragraphs are separated based on the logic-based component such as optimization, sampling, PF, etc. 

Optimization-based trajectory generation methods can also be incorporated into the learning-based motion planning framework.  In~\cite{hegedus2019motion} the computational time of the optimization method is reduced by developing an artificial neural network (ANN) that is trained to learn the outputs of the optimization algorithm. The optimization algorithm supervises the output of the ANN before generating the final trajectory to ensure that the safety constraints are met. Similarly, Alexandru et al. in~\cite{tuatulea2020design} used the obtained trajectories from a nonlinear MPC along with the corresponding inputs and trained an alternative ANN to address the high computational cost challenge of the optimization-based algorithm. In another study, the authors in~\cite{sun2018fast} addressed the challenge of computational complexity by proposing a hierarchical layered structure in which the first layer consists of a neural network trained by MPC, and the next layer is responsible for guaranteeing the feasibility of the planned trajectory. The hybrid framework has been also developed by learning-based TDM and optimization-based TG~\cite{yang2022hybrid}. In their publication, the interaction between the EV and human-driven vehicles is captured during TDM through a learning-based method, while the subsequent TG process is in charge of generating the optimized trajectory to meet the upstream decision within a mixed traffic flow context.

\begin{figure*}[t]
\centering
\includegraphics[width=1\linewidth]{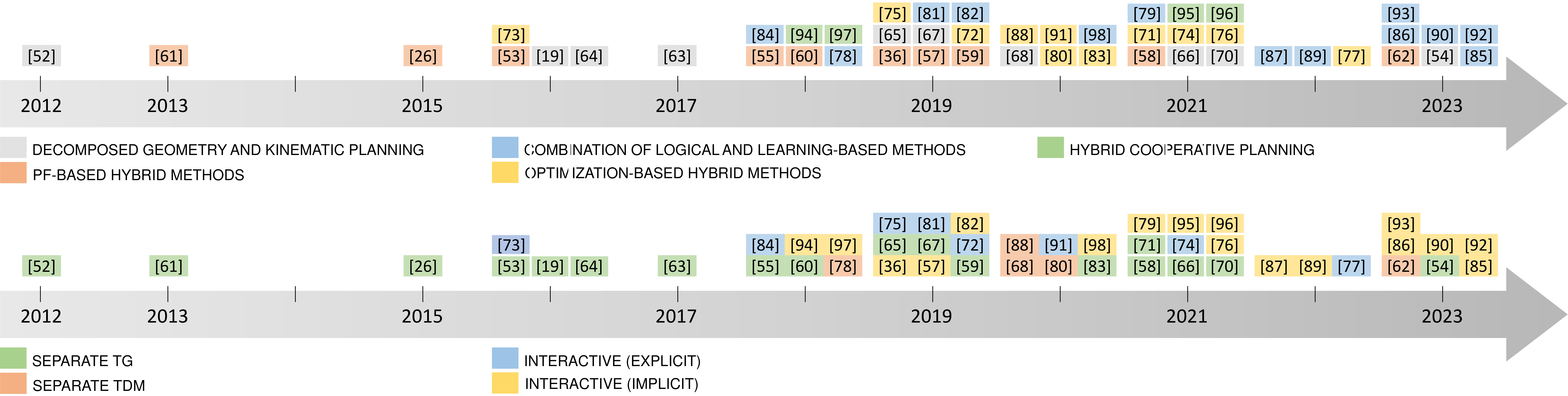}
\caption{Timeline of different types of hybrid motion planning methods based on the hybrid components(top), and architecture of the motion planning (bottom).}
\label{fig: timeline}
\centering
\end{figure*}

The learning-based approach has also been used to improve the performance of sampling-based motion planning methods. Zhang et al.~\cite{zhang2022integrating} developed a hybrid planning framework in which the sampling points are dynamically adjusted based on the driving context, i.e., the actor behaviours and the road layout. Likewise, in a separate investigation~\cite{wen2023tofg}, improvements are made to sampling-based motion planning through a two-step process. Initially, an attention-based neural network is utilized in conjunction with the temporal occupancy flow concept~\cite{mahjourian2022occupancy} to capture the interaction between various semantic information and the future state of dynamic actors in several driving scenarios. Subsequently, trajectory samples are generated based on the attention map obtained in the preceding step. The findings presented in their research demonstrate that this approach effectively addresses the motion planning challenge across diverse driving scenarios, all without requiring pre-established knowledge about the driving context. A review of similar hybrid approaches in robotic motion planning has been published in~\cite{qureshi2020motion}. In another research~\cite{yu2021reducing}, Graph Neural Networks have been used to improve the sampling-based motion planners by reducing the collision checks in the sampling process. 

Recently, prediction and motion planning have been jointly addressed by incorporating optimization-based methods into a hybrid framework along with learning-based approaches~\cite{hagedorn2023rethinking}. In~\cite{huang2023differentiable} and~\cite{sun2023get}, the authors proposed an integrated prediction-planning attention-based network in which a differentiable cost function is designed to generate the final trajectory, while the input signals are the longitudinal acceleration and steering angle. In order to enable learning through backpropagation the kinematic bicycle model must be linearized.  Notably, this framework stands out because the cost function itself is fine-tuned through the training process, as the motion planning cost function is integrated into the neural network's loss function. Imitation learning (IL) prediction methods could be used for human-like motion planning or driving \cite{wen2023tofg}. In a recent study released by Liu et al.~\cite{liu2023occupancy}, a transformer-based DNN was employed to tackle the simultaneous tasks of joint motion prediction for other road users and motion planning for the EV. The authors utilized IL-based learning to identify the driving mode from a set of pre-defined modes. Subsequently, they refined the chosen mode through open-loop optimization within a finite horizon for calculating the final planned trajectory. Also, there are further studies~\cite{sormoli2023novel, wen2023tofg} integrating IL with nature-inspired logics such as fluid flow simulation \cite{sulkowski2022autonomous} to guide neural networks capturing the interactions between actors and subsequently mitigating the training effort.

The combination of PFs and learning-based methods is another example in this category of hybrid motion planners. Li et al. in~\cite{li2018humanlike} trained a convolution neural network (CNN) for achieving human-like decision-making. The input to the CNN is a bird's eye view representation of the perceived environment, while the outputs are the speed and steering angle of the EV. The final outputs are calculated by adding in the last layer of the CNN the weighted repulsive artificial forces applied to the EV by other vehicles in lateral and longitudinal directions. Their study showed that PFs can effectively reduce the number of layers in the CNN, while successfully imitating human driving behaviour and ensuring safety. Wang et al. in~\cite{wang2021imitation} used the concept of artificial PF to reduce the complexity associated with the learning phase. They extracted an intention potential map (see also~\cite{ma2020deepgoal}) from the route (obtained by a high-level planner) and the front view RGB image. Subsequently, they combined this PF with an obstacle potential map obtained directly from a LiDAR point cloud to construct an artificial PF-like map. In the next step, the resulting potential map is utilized as input to a CNN to plan the trajectory.

The combination of reinforcement learning (RL) motion planning algorithms with classical methods has been also introduced recently.  Wang et al. in~\cite{wang2019quadratic} developed a hybrid control system for AVs to tackle the challenge of discrete action space used in reinforcement learning methods. They used a combination of Q-learning and traditional proportional–integral–derivative (PID) control methods, instead of training a neural network with hundreds of thousands of neurons, which is a time-consuming task that requires significant computational power. Furthermore, they introduced another neural network to calculate the tuning parameters of the PID controller responsible for generating the final continuous actions (lateral/longitudinal). In this way, the motion planner could adapt to various driving scenarios. In another study~\cite{wang2020learning}, the reward is designed based on a sampling-based trajectory planner. Therefore, this leads to a policy that optimizes both behaviour and motion at the same time and also produces smooth trajectories. Fuzzy logic is another tool that can improve the overall performance of the motion planning module when combined with learning-based methods such as artificial neural networks or RL. Chen et al.~\cite{chen2020conditional} proposed a hybrid end-to-end motion planning framework based on CNNs. After extracting spatial and temporal features from camera images using a CNN and a long short-term memory (LSTM) network, respectively, they used these features as input to a fully connected neural network to calculate the control commands, i.e., steering angle and acceleration/deceleration. However, in the last layer, instead of using directly the control commands, they calculated two sets of fuzzy parameters and finally obtained the control commands by using the maximum defuzzification. To do that, they formulated the dependency of two output commands by fuzzy logic, instead of including them in the neural network and making the learning process more complicated. For example, a high turning rate and high acceleration should not happen simultaneously. This approach leads to more stable and smooth control commands. 

To summarise, during offline training, learning-based methods can supervise a motion planner that uses a traditional optimization-based approach to achieve near-optimal performance. At the same time, PFs can be employed to reduce the training complexity, because they are very efficient in describing complex driving scenarios/environments in a simplified manner.   
\\

\subsubsection{Hybrid Cooperative Planning} The methods reviewed in the preceding sections for TG and TDM were based on the EV's standalone hardware and software capabilities. However, the advent of wireless vehicle-to-everything (V2X) communication technologies such as dedicated short-range communication (DSRC), ITS-G5 and cellular V2X (C-V2X) would enable widespread connectivity between vehicles or between vehicles and the infrastructure or the cloud.  The influences of these communication technologies on ADS have been broadly studied in~\cite{talebpour2016influence,liu2018modeling,Hazim2023}. Specifically, the perception of the EV in challenging driving situations with limited field-of-view or occlusions could be enhanced by fusing its onboard sensor data with off-board information received over a V2X communication system~\cite{arnold2019survey,arnold2022}. The hybrid motion planning methods that take advantage of the shared information to further enhance driving efficiency and improve the traffic flow are reviewed in this subsection. 

The motion control of connected and automated vehicles is divided into two main categories: individual and batch control. Individual control is similar to the standalone motion control, but the trajectories and/or the decisions of other road participants are also provided as inputs to the motion planner of the EV. For example, instead of predicting the future intentions of other vehicles, the EV can obtain their planned and desired manoeuvres and trajectories using, for instance, the manoeuvre coordination service (MCS) messages received over V2X~\cite{MCM}. In a similar fashion, the EV can receive the future intentions of vulnerable road users over cellular connectivity.  In batch control, there is a central or high-level processing unit that considers motion planning for a fleet of vehicles rather than a single vehicle. 

For individual cooperative motion control, the performance of PFs, optimization-based methods~\cite{eilbrecht2018optimization} (especially MPC) and their combination have been already investigated in~\cite{viana2021comparison}. The only difference as compared to non-cooperative motion planning is the higher accuracy attained in predicting the states of other vehicles. For instance, the terms used for penalizing the collisions between the EV and other obstacles in the cost function of the MPC algorithm become more realistic, and accordingly, the planned trajectory/decision would be more reliable. On the contrary, in the case of batch control such as traffic control and platooning, the interaction between vehicles has to be modelled as well, in addition to the dynamic model of each road participant. The hybrid framework has been also used in the case of batch control to improve the performance of the motion planning system. Hidalgo et al.~\cite{hidalgo2021platoon} designed a hybrid TG algorithm based on MPC and parametric curve algorithm responsible for lane changing tasks in platoon merging along with feedback/feedforward controller for longitudinal control in order to guarantee real-time performance. In another study, Huang et al.~\cite{huang2018path} used a combination of artificial PF and MPC algorithms to develop a multi-vehicle cooperative platoon control.

It is important to note that there is also another type of hybrid framework used in cooperative vehicle motion planning and control which is known as \textit{hybrid automata}. Unlike the previous hybrid methods where the combination of different methods is used for designing a single module, e.g., TG or TDM, hybrid automata are used to combine both modules of TG (continuous) with TDM (discrete). Therefore, the interactions between these modules (signals \textbf{B1} and \textbf{B2} in Fig.~\ref{fig:cont_layout}) are considered in the model and the overall control performance increases accordingly. For instance, in~\cite{huang2018path,ma2020novel} both discrete manoeuvre switches and continuous motion control of the cooperating vehicles have been formulated by the hybrid automaton model. However, there are other works that address the same challenge by fusing decision-making and TG tasks implicitly~\cite{gu2016automated}.
\\
\subsubsection*{Categorizing based on TDM and TG interactions} The hybrid motion planning methods reviewed so far are also summarized in Table~\ref{table:hybrid}, where one can easily retrieve the combined elements of each method along with the addressed motion planning challenges. Before comparing the various hybrid methods that have so far appeared in the literature, we would also like to note that these methods can be further categorized according to the interaction between the underlying TDM and TG processes. This can further clarify the motion planning challenges targeted by each hybrid approach. As listed in Table~\ref{table:interaction}, while some hybrid motion planners (\textit{separate}) focus either on the TG or on the TDM process, there are some hybrid approaches (\textit{interactive}) that lead to a combined TG and TDM algorithm with either implicit or explicit interaction between the two processes. The categorization of the existing literature in hybrid motion planning methods for AVs in terms of (i) classical methods as building block elements, (ii) addressed challenges, and (iii) TDM and TG interactions, is another contribution of this review article that will also help in the identification of research gaps.

\section{Discussion and open challenges}
\label{compare}
Although the reviewed hybrid motion planning methods have not been evaluated or compared using the same dataset, in this section, their performances are discussed and assessed based on the challenge(s) they aimed to address. Moreover, research gaps and potential directions for future work are highlighted in the latter part of this section.

\subsection{Performance Assessment and Comparison}
%The evaluation of the hybrid approaches is based on their ability to address the motion planning challenges  detailed in Section~\ref{hybrid}. 
The classification of approximately $50$ hybrid motion planning studies under the five groups described in Section~\ref{sec:hybrid} and the evaluation of their performances against the five challenges (\textit{``a"} to \textit{``e"}) are summarized in Table~\ref{table:hybrid}. % (``Challenges" columns).
One can see over there that for the time being there is no hybrid method addressing more than three challenges. The most popular category is the  \textit{``decoupled geometric and kinematic planning"} that reduces the high computational load of motion planning by separately designing the path (using parametric curves) and the kinematic features of the trajectory. In spite of enabling real-time applications, the planned trajectory is likely to be suboptimal, and furthermore, due to the fact that the non-holonomic constraints of the vehicle are not explicitly considered, its feasibility is not guaranteed either.

The \textit{``hybrid methods using potential fields"} appear in various forms in the literature, but researchers, mainly, have incorporated PFs in \textit{``PF-optimization"} and \textit{``PF-learning"} hybrid structures to improve the performance of TG  and/or TDM. In the \textit{``PF-optimization"} hybrid framework, a PF is generated by abstracting the geometry of the drivable area, and the relative distance/velocity of other road users, in addition to other semantic information such as lane markings used to quantify driving safety and comfort as a cost function. Combined with receding horizon optimization methods including the dynamic model of the vehicle, a PF enables reactive and feasible motion planning. Although this hybrid approach could be used for TG (and even TDM) for a wide range of scenarios, since the semantic information of the driving context is highly abstracted into a scalar cost value, it needs fine-tuning to adapt to new scenarios. In the \textit{``PF-learning"} hybrid structure, a PF is used for modifying either the input or the output of the ANN. In the former case, the PF is used to interpret the semantic information as a risk map to reduce the complexity of the end-to-end algorithms. In the latter case, the PF is used to ensure that the control actions (outputs) are safe and reasonable by encoding the artificial repulsive force at the output of the neural network. Similarly, \textit{``Fuzzy logic"} has been also used to reduce the complexity of the learning-based approaches by encoding the correlations in the outputs of the neural network to achieve smoother, more logical, and safer outputs. Finally, the main objective of \textit{``learning-based and logical combination methods"} is to improve explainability and facilitate debugging while providing general-purpose motion planning that applies to several driving scenarios, only by changing the training data without major changes in the structure of the neural network. 

The hybrid approaches discussed above have been also used for \textit{``cooperative control"}. The \textit{``hybrid automata"} is a special framework used in cooperative motion planning with the scope to address the interaction between TDM and TG, which are discrete and continuous systems (processes), respectively. Although some studies report a promising performance for the \textit{``hybrid automata"} approach regarding motion planning of distributed systems like platooning, the number of manoeuvres that this approach can handle is rather limited, and predefined (primitive) motions used in this method put up a challenge on the feasibility of the planned trajectories. 
  
\subsection{Research Gaps and Opportunities For Future Studies}
According to the works reviewed in this survey, a hybrid framework is a promising approach to overcome several challenges associated with the motion planning problems of AVs. The vehicle-based and environment-based challenges are mostly covered in the existing literature, however, the safety and especially uncertainty-related challenges are yet to be studied in-depth. Table~\ref{table:hybrid} demonstrates this fact as well, where one can see that challenge ``e" (uncertainty-based challenges) has not been the main objective of research studies so far. To highlight the impact of various uncertainties on motion planning, consider, for example, that the reliability of perception and localisation due to various system/sensor faults is closely related to the fail-safe control of AVs. Therefore further research focusing on (hybrid) motion planning methods that are resilient to imperfections in perception and localisation is a promising direction to follow. 

Moreover, since the existing hybrid motion planning methods are applicable only to a limited number of scenarios, future studies could be focused on developing a general framework, in which the interactions between various TDM  and TG algorithms are considered for a safe transition from one mode or manoeuvre to another. This is also likely to reduce the frequency of handover (or fallbacks) requests to the safety driver, which translates to higher SAE levels of autonomy. While several studies attempt to design a general-purpose motion planner and cover challenge ``b" (driving context challenges), only two of them, according to Table~II, consider at the same time the feasibility of the generated trajectories, i.e., challenge ``a"  (vehicle dynamic and feasibility). Therefore, the development of general-purpose motion planners which also address the remaining challenges remains open. 

Finally, another research direction worth pursuing stems from the strong coupling between motion planning and the behaviour/intention prediction of other road users. Decision-making should take into account the predicted intentions of other road users, however, their intentions can dynamically change given the decisions/behaviour of the EV and vice versa. It is expected that this coupling would further increase the real-time implementation requirements of motion planning and control algorithms (challenge ``c"), especially in high-speed environments, such as motorway merging and motorway chauffeur. Therefore further research on hybrid approaches with the aim to reduce their computational complexity would be highly beneficial for such scenarios. %The interactions between the prediction module and motion planning must be properly incorporated into the tactical decision-making and trajectory generation methods and also   

\section{Summary and Conclusions}
\label{conclusion}
This review shows that the current trend for designing motion planning algorithms for AVs is based upon a hybrid framework that combines various traditional algorithms together. We have defined four categories of hybrid motion planners which encompasses, to the best of our knowledge, all hybrid methods in the existing literature, namely, (i) \textit{decomposed geometry and kinematic planning}, (ii) \textit{hybrids using potential fields}, (iii) \textit{optimization-based hybrids}, and (iv) \textit{combinations of logical and learning-based methods}. We have also defined, as a separate category, the \textit{hybrid cooperative motion planning}, which uses either V2X communication to assist individual vehicles in motion planning or batch control for fleets of vehicles.  The performance assessment of around $50$ hybrid methods falling under the above categories shows that combining two or more traditional motion planning methods together is a promising approach that can help cancel out the shortcomings of each separate method, without sacrificing the advantages of its components. According to the timeline presented in Fig.~\ref{fig: timeline} (top), the prevailing approach during the last two years has been the {\textit{combination of logical and learning-based methods}}. With the ongoing advances in computing capabilities and machine learning we expect this trend to continue and grow.  

Furthermore, we have identified the following key challenges for motion planning and used them as the metric to compare different  methods: (a) \textit{Vehicle’s dynamics and feasibility}, (b) \textit{driving context}, (c) \textit{real-time implementation}, (d) \textit{safety-based}, and (e) \textit{uncertainty-based} challenges. Our review shows that the main focus of existing hybrid methods is to properly model the {\textit{driving environment}} and reduce the {\textit{computational time}}, in order to enable reliable and safe motion planning that can operate in real-time. Nevertheless, there are still remaining issues to overcome, such as {\textit{uncertainties}} in perception and localisation that can affect the {\textit{safety}} of generated trajectories. Hybrid methods addressing \textit{uncertainty-based} challenges are under-represented in the available literature. For example, confidence intervals associated with the output of the perception/localisation modules could be incorporated into the motion planner to ensure safety, which is a promising direction for future work. 

Finally, another contribution of this review is to categorize the hybrid techniques based on the interactions between the tactical decision-making (TDM) and trajectory generation (TG) modules. 
This has revealed a lack of generality in TDM and TG methods with respect to the driving scenario, which is a recommended direction for further research. General-purpose motion planners can help reduce the frequency of fallbacks to the safety driver and enable higher SAE levels of autonomy. The current trend is an \textit{interactive implicit design} between TDM and TG, see Fig.~\ref{fig: timeline} (bottom), which is expected to continue, cultivating a comprehensive interaction between these two processes to cover a wide range of possible driving scenarios. We believe that this survey paper will spark more research activities on motion planning for modular automated driving systems, and help researchers and industries to better position their work in terms of the building block algorithms for hybrid motion planning, the addressed challenges, and the interactions between the underlying TDM and TG processes. % which are often used interchangeably in the literature.

\bibliographystyle{myIEEEtran}
\bibliography{IEEEabrv,bibliography}

\vskip 0pt plus -1fil

\begin{IEEEbiography}[{\includegraphics[width=1in,height=1.25in,clip,keepaspectratio]{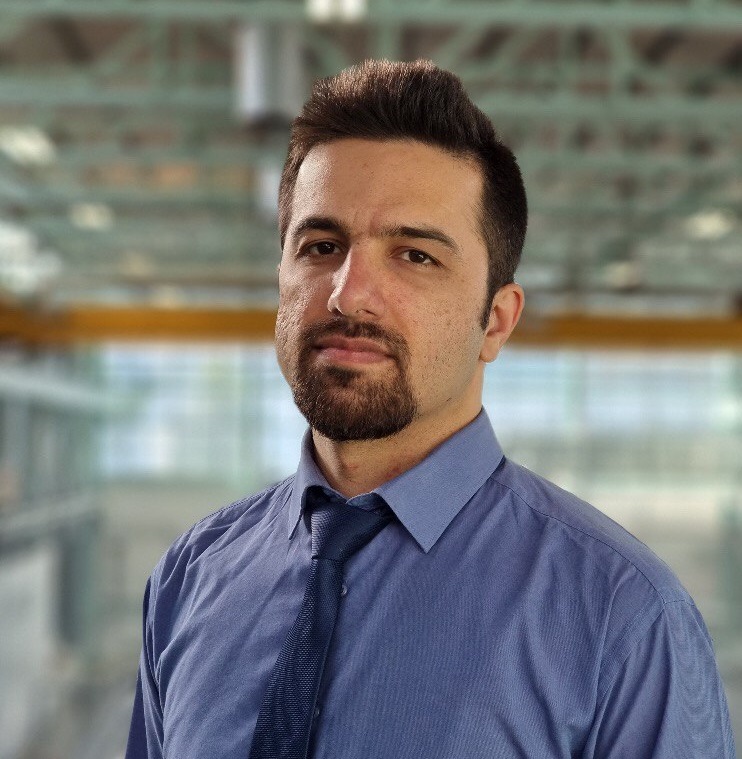}}]{Mohammadreza Alipour Sormoli}
received the M.Sc. degree from the Amirkabir University of Technology (Tehran Polytechnic) in 2017. worked as a research assistant at Koc University and is currently working toward the PhD degree in the field of autonomous driving technology at the University of Warwick (WMG). His research interests include robotics, mechatronics, control and dynamics of autonomous systems.
\end{IEEEbiography}

\begin{IEEEbiography}[{\includegraphics[width=1in,height=1.25in,clip,keepaspectratio]{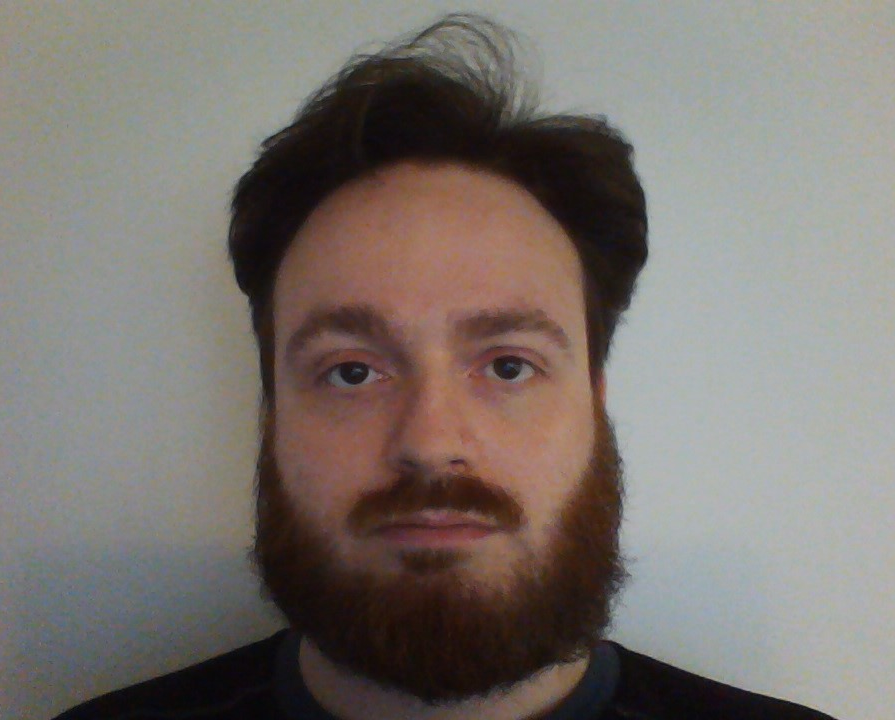}}]{Konstaninos Koufos} obtained the 5-year Diploma in Electrical \& Computer Engineering from Aristotle University of Thessaloniki, Greece, in 2003 and the M.Sc. and D.Sc. in Radio Communications from Aalto University, Finland, in 2007 and 2013. He worked as a post-doctoral researcher on 5G wireless networks at Aalto University, as a Senior Research Associate in Spatially Embedded Networks at the School of Mathematics at the University of Bristol, UK, and as a Senior Research Fellow on Cooperative Autonomy within the Warwick Manufacturing Group (WMG) in the University of Warwick, UK. He is an Assistant Professor on Future Mobility Technology at WMG, working on perception, motion planning, and fail-safe control of connected and automated vehicles. His research interests also include stochastic geometry and performance modelling of mobile wireless networks.
\end{IEEEbiography}

\begin{IEEEbiography}[{\includegraphics[width=1in,height=1.25in,clip,keepaspectratio]{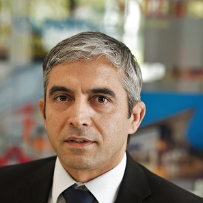}}]{Mehrdad Dianati}
(Senior Member, IEEE) is a professor of connected and cooperative autonomous vehicles at WMG, the University of Warwick and the School of EEECS at the Queen's University of Belfast. He has been involved in a number of national and international projects as the project leader and the work-package leader in recent years. Prior to academia, he worked in the industry for more than nine years as a Senior Software/Hardware Developer and the Director of Research and Development. He frequently provides voluntary services to the research community in various editorial roles; for example, he has served as an Associate Editor for the IEEE Transactions On Vehicular Technology.  He is the Field Chief Editor of Frontiers in Future Transportation.
\end{IEEEbiography}

\begin{IEEEbiography}[{\includegraphics[width=1in,height=1.25in,clip,keepaspectratio]{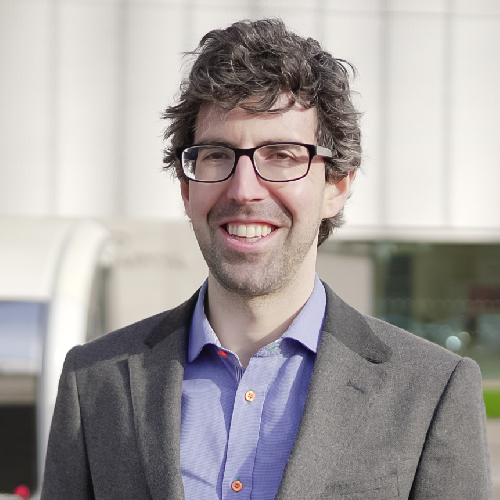}}]{Roger Woodman}
is an Assistant Professor and Human Factors research lead at WMG, University of Warwick. He received his PhD from Bristol Robotics Laboratory and has more than 20 years of experience working in industry and academia. Among his research interests, are trust and acceptance of new technology with a focus on self-driving vehicles, shared mobility, and human-machine interfaces. He has several scientific papers published in the field of connected and autonomous vehicles. He lectures on the topic of Human Factors of Future Mobility and is the Co-director of the Centre for Doctoral Training, training doctoral researchers in the areas of intelligent and electrified mobility systems.
\end{IEEEbiography}

\end{document}